\documentclass[oneside]{article}

\usepackage[utf8]{inputenc}
\usepackage{amsmath}
\usepackage{amssymb}
\usepackage[margin=1in]{geometry}
\usepackage{mathtools}
\usepackage{graphicx}
\usepackage{url}
\usepackage{algorithm}
\usepackage{algpseudocode}
\usepackage{hyperref}
\usepackage[capitalize,poorman]{cleveref}
\usepackage{verbatim}
\usepackage{xfrac}
\usepackage{caption}
\usepackage{subcaption}
\usepackage{pifont}
\usepackage{multirow}
\usepackage{booktabs}

\begin{document}

\title{Graph Partitioning and Sparse Matrix Ordering using Reinforcement Learning and Graph Neural Networks}

\author{Alice Gatti\thanks{Computational Research Division, Lawrence Berkeley National Laboratory} \and Zhixiong Hu\thanks{University of California, Santa Cruz} \and Tess Smidt\footnotemark[1] \and Esmond G.~Ng\footnotemark[1] \and Pieter Ghysels\footnotemark[1]}

\date{\today}

\maketitle

\begin{abstract}
  We present a novel method for graph partitioning, based on reinforcement learning and graph convolutional neural networks. Our approach is to recursively partition coarser representations of a given graph. The neural network is implemented using SAGE graph convolution layers, and trained using an advantage actor critic (A2C) agent. We present two variants, one for finding an edge separator that minimizes the normalized cut or quotient cut, and one that finds a small vertex separator. The vertex separators are then used to construct a nested dissection ordering to permute a sparse matrix so that its triangular factorization will incur less fill-in.  The partitioning quality is compared with partitions obtained using METIS and SCOTCH, and the nested dissection ordering is evaluated in the sparse solver SuperLU. Our results show that the proposed method achieves similar partitioning quality as METIS and SCOTCH. Furthermore, the method generalizes across different classes of graphs, and works well on a variety of graphs from the SuiteSparse sparse matrix collection.
\end{abstract}

\section{Introduction\label{sec:intro}}
The problem of partitioning a graph into approximately equal sized subgraphs while minimizing the number of cut edges is an NP-complete problem \cite{npProblemsMichaelJohnson}. Practical graph partitioning algorithms for large scale problems are based on heuristics which give approximate solutions. In general it is not known how far these approximations are from the optimal solution. Moreover, many of the heuristic algorithms are inherently sequential in nature and do not exploit current high-performance computing hardware efficiently. Heuristic algorithms exhibit very irregular memory access patterns -- leading to low memory bandwidth usage because of wasted cache lines and branch prediction misses -- and perform mostly integer manipulations. Deep learning is an expressive and flexible algorithmic framework that can run efficiently on  modern hardware, making it a powerful tool for tackling the problem of graph partitioning from a new perspective. 
%\newpage\noindent

In recent years there has been a growing interest in using machine learning techniques to find approximate solutions of NP problems, like the travelling salesman problem, knapsack problem, vertex cover etc. Many of the problems in this class are naturally formulated as combinatorial optimization problems over graphs, and powerful learning tools to handle this type of data are graph neural networks \cite{brunaBeyoindEuclideanData,Hamilton2017RepresentationLO,Battaglia2018RelationalIB}, which are an extension of usual deep neural networks to non-Euclidean data such as graphs. One of the key ingredients of graph neural networks is graph convolution, that generalizes convolution over grids to graph data structures. This generalization is often called message passing and it allows to aggregate local information on the neighbors and propagate it on the graph. This scheme has been integrated in many different convolutional layers such as GCN \cite{gcn}, SAGE \cite{hamilton2017inductive} and GAT \cite{attention2018}, in which message passing is combined with the attention mechanism. Several machine learning methods including graph neural networks have been proposed to approximate the solutions of the travelling salesman problem \cite{zheng2021travsales, neuralCOgraphs2017, LearningTS2019}, knapsack problem \cite{knapsack2020}, vertex cover and maximum cut \cite{LearningCO2017}. These works propose supervised and unsupervised methods together with reinforcement learning. There are also works focused on the graph partitioning problem. \cite{karalias2021erdos} combines a probabilistic method with graph neural networks, but it has not been tested on big graphs. \cite{nazi2019gap} presents an unsupervised deep learning method that seems promising from the performance point of view, but it requires graphs with many features and it was tested on small graphs as well. More generally, deep learning (DL) methods for graphs are currently an area of active development with several powerful tools emerging from the community such as PyTorch geometric~\cite{Fey_Lenssen_2019} and the Graph Nets library~\cite{graph_nets} for TensorFlow~\cite{tensorflow2015-whitepaper}.
Much of the recent DL work has been targeted at GPUs and TPUs, which are becoming integrated components in modern high performance computing architectures.

We present a novel approach for graph partitioning based on deep reinforcement learning and graph convolutional layers. The proposed method refines a partition which is computed from a coarser representation of the graph. In the refinement procedure, an agent, implemented using SAGE graph convolutional layers from~\cite{hamilton2017inductive}, moves nodes from one partition to the other in order to improve the normalized cut of the partitioning. The agent is trained using the advantage actor critic reinforcement learning algorithm. A similar deep reinforcement learning algorithm is also used to partition the coarsest graph in the multilevel scheme. This multilevel graph bisection scheme is then modified in a subsequent section to directly find a vertex separator. Vertex separators are used in the nested dissection ordering algorithm, which is a heuristic to reduce the fill-in in sparse direct solvers. We implement a nested dissection sparse matrix ordering algorithm where the vertex separators are computed using the proposed multilevel deep reinforcement learning based algorithm, and we evaluate the quality of this ordering using the SuperLU~\cite{li2005overview} sparse solver.

% \todo[inline]{discuss our new approach in more detail: reward function, learning algorithm (DQN/A2C?), ...}
% \todo[inline]{Discuss other deep learning approaches from literature here??}

% Since many graph problems are NP-complete, or many NP-complete problems can be formulated as graph problems, we believe our method could find applications other than graph partitioning; or future developments in graph partitioning could be influenced by progress on related graph problems.

We use the following notation: a graph $G=(V,E)$ has a set of vertices $V = \{v_1, \dots, v_n \}$ connected by edges $E = \{e_{ij}  = (v_i, v_j) \, |\, v_i, v_j \in V\}$. The total number of vertices, also called nodes, is denoted by $|V| = n$. We define the degree of a vertex as $\text{deg}(v)=\sum_{(v,w)\in E}1$ and throughout the paper we will assume, without loss of generality, that all graphs are undirected and fully connected.
%The degree of a vertex is the number of edges connecting it.
%We assume graphs are undirected and, without loss of generality, each graph has a single connected component.

\begin{figure}
    \centering
    \begin{subfigure}[b]{0.4\textwidth}
      \centering
      \includegraphics[scale=1]{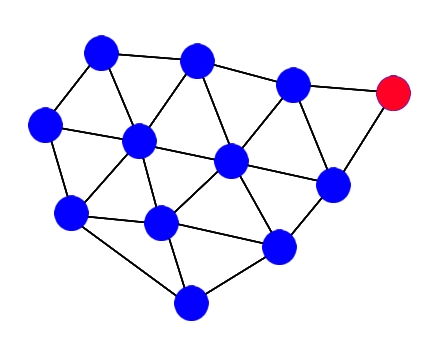}
    \caption{Minimum cut}
    \label{fig:graphCutBad}
    \end{subfigure}
   \begin{subfigure}[b]{0.4\textwidth}
      \centering
      \includegraphics[scale=1]{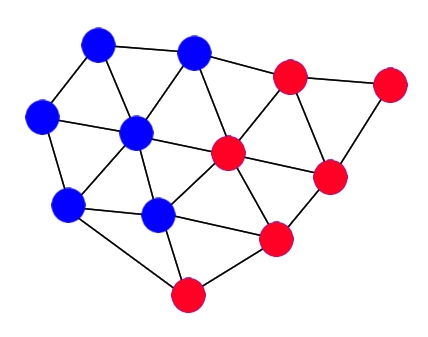}
      \caption{Minimum normalized cut}
      \label{fig:graphCutGood}
   \end{subfigure}
   \caption{Graph partitioned into different ways. The left partition has minimum cut, but the sizes of the partitions are highly unbalanced, since all nodes except one are in the blue partition. The right partition has higher cut, but the partitions have volumes $22$ and $24$, hence they are approximately balanced.}
   \label{fig:minimumCut}
\end{figure}

% \subsection{Graph Partitioning for Scientific Computing\label{ssec:graph_partitioning}}
Graph partitioning is the problem of grouping the nodes of a graph $G = (V,E)$ into non-empty disjoint subsets. For graph bisection, i.e., partitioning the graph in two sub-graphs $G_A = (V_A,E_A)$ and $G_B = (V_B,E_B)$, such that $V_A \cap V_B = \emptyset$ and $V = V_A \cup V_B$. Often $k$-way graph partitioning is implemented using recursive bisection. However, this does not always guarantee well-balanced partitions, see Figure \ref{fig:minimumCut}. The goal of partitioning is often to minimize the number of edges between the partitions, i.e., the cut, while keeping the sizes -- either cardinality or volume, i.e., the sum of the node degrees -- of the partitions balanced. %\todo{Do we need to define ``volume'' here?} 

A popular heuristic for graph partitioning is the Kernighan-Lin (KL)~\cite{kernighan1970efficient} method. However, this has a cost of $O(|V|^2d)$, where $d$ is the maximum node degree in $G$. A variation on KL, the Fiduccia-Mattheyses~\cite{fiduccia1982linear} algorithm, reduces this cost to $O(|E|)$. The Fiedler vector is the eigenvector corresponding to the smallest non-zero eigenvalue $\lambda_2$ of the graph Laplacian $L(G) = D - A$, where $D$ is a diagonal matrix with the node degrees, and $A$ is the adjacency matrix of $G$. In spectral partitioning~\cite{pothen1990partitioning,simon1991partitioning} the values of this vector are used for partitioning. The Cheeger bound~\cite{cheeger1969,chung1996spectral} guarantees that spectral bisection provides partitions with nearly optimal conductance (the ratio between the number of cut edges and the volume of the smallest part). Spectral methods make use of global information of the graph, while combinatorial algorithms like KL and FM rely on local information.

\begin{figure}[t]
    \centering
    \includegraphics[scale=0.85]{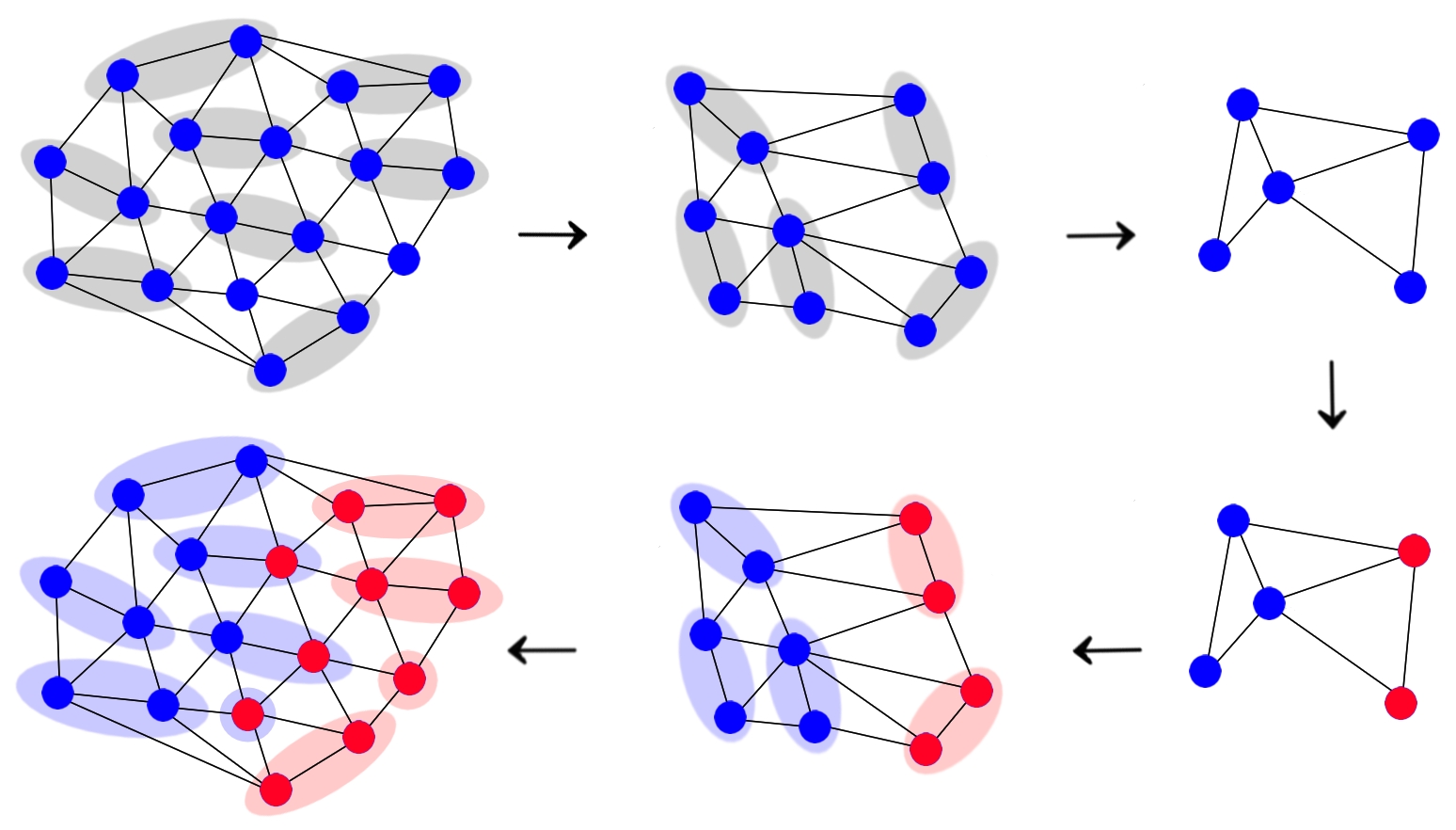}
    \caption{Multilevel approach for graph partitioning. On the top of the picture the graph is coarsened recursively, until the coarsest graph as few vertices. The coarsening is made by matching vertices included in the same gray ellipse. Then, on the bottom part of the figure, the coarsest graph is partitioned and it is interpolated back by one level, where the partition is further refined. The interpolation/refinement continues up to the initial graph. The vertices included in the red or blue ellipses have been interpolated from a red or blue node respectively.}
    \label{fig:coarseningRefining}
\end{figure}

State-of-the-art graph partitioning codes typically use a multilevel approach illustrated in Figure \ref{fig:coarseningRefining}: first the graph is coarsened, then a much smaller graph is partitioned and then this partitioning is interpolated back to the finer graph where it can be refined. Different algorithms can be used for the coarse partitioning and for the refinement.

Although the KL and FM algorithms are widely used in practice, they have limited parallelism and have irregular memory access patterns and lots of integer manipulations, making them inefficient on modern hardware. Spectral graph methods on the other hand, are based on eigenvalue solvers and can take advantage of the huge computational power of modern CPUs, including SIMD units, and GPUs. Still, spectral methods can be expensive, as the number of iterations for the eigensolver, for instance Lanczos~\cite{lanczos1950iteration}, Rayleigh quotient iteration~\cite{trefethen1997numerical} or LOBPCG \cite{knyazev2001toward}, can be large. Popular graph partitioning libraries are METIS~\cite{karypis1998fast} and SCOTCH \cite{pellegrini:DSP:2009:2091}, and their parallel versions ParMETIS~\cite{karypis1999parallel} and PT-SCOTCH~\cite{chevalier2008pt}, as well as Zoltan(2)~\cite{devine2002zoltan}. 

%Partitioning of graphs, meshes or sparse matrices is an important pre-processing step in many high-performance parallel computing simulations, with the goal of distributing work evenly over compute nodes while minimizing communication in, for instance, Krylov iterative solvers and preconditioners like algebraic multigrid (AMG)~\cite{falgout2002hypre}, block Jacobi and domain decomposition~\cite{smith2004domain}. For instance the widely used scientific computing codes PETSc~\cite{petsc-user-ref}, Trilinos~\cite{heroux2005overview} and the MFEM finite element library~\cite{anderson2019mfem} all use METIS for graph partitioning.
Graphs arise frequently in scientific computing.  For example, a mesh used in the discretization of partial differential equations can be considered a graph.  Sparse matrices are another example; their sparsity structures can be represented by graphs as well.
Partitioning of graphs is an important pre-processing step in high-performance parallel computing, with the goal of distributing the computational work evenly over the compute nodes while minimizing communication in, for instance, Krylov iterative solvers and preconditioners like algebraic multigrid (AMG)~\cite{falgout2002hypre}, block Jacobi and domain decomposition~\cite{smith2004domain}. For instance the widely used scientific computing codes PETSc~\cite{petsc-user-ref}, Trilinos~\cite{heroux2005overview} and the MFEM finite element library~\cite{anderson2019mfem} all use METIS for graph partitioning.

The remainder of the paper is organized as follows. Section \ref{sec:a2c} briefly introduces reinforcement learning, in particular the distributed advantage actor critic (DA2C) training algorithm. In Section \ref{sec:edge_separator} we present an algorithm to compute a minimal edge separator using deep reinforcement learning within a multilevel framework. Section \ref{sec:vertex_separator} shows a variation of this algorithm presented to compute a vertex separator instead of an edge separator. The vertex separator algorithm is used in Section \ref{sec:nested_dissection} to construct a nested dissection sparse matrix ordering. 
We conclude the paper with a summary and outlook in Section \ref{sec:conclusion}. All the codes are made available at the GitHub page \url{https://github.com/alga-hopf/drl-graph-partitioning}.

%\todo[inline]{outline}
% Section~\label{sec:background} introduces our notation and describes problem of graph partitioning, its many applications and the current state of the art. In Section~\ref{sec:reinforcement_learning} we give a brief overview of reinforcement learning techniques. Section~\ref{sec:model} describes our novel reinforcement learning based approach to graph partitioning. Then, in Section~\ref{sec:evaluation} we present detailed experimental evaluation of our new method and a comparison with existing widely used alternatives. Finally, Section~\ref{sec:conclusions} concludes this paper with a summary and outlook.

\section{Advantage Actor Critic\label{sec:a2c}}

% https://stats.stackexchange.com/questions/250943/what-is-the-difference-between-episode-and-epoch-in-deep-q-learning

Reinforcement learning~\cite{suttonrl2018} is an extremely flexible framework for training an agent, interacting with an environment, to maximize its cumulative reward. In particular, the agent acts on instances of the environment, called states, by taking actions following a certain policy $\pi$, that determine the transition to another state. We denote by $\mathcal{S}$ the set of states of the environment, $\mathcal{A}(s)$ the set of actions
%of
that the agent can take in state $s$ and by $r : \mathcal{S} \times \mathcal{A} \to \mathbb{R}$ the reward function. A reinforcement learning problem may be described by a (finite) Markov decision process (MDP) $(s_t, a_t, r_{t+1})_{t\in [0,T-1]}$,
% \todo{Esmond:  Not sure if the last subscript ($t$) is needed.}
where $s_t \in \mathcal{S}$, $a_t\in\mathcal{A}(s_t)$ are the state and action at time $t$, while $r_{t+1} := r(s_t, a_t)$ is the reward received after action $a_t$ is performed. The time $t$ runs from $0$ to $T$, the time step at which the episode ends. Since the process is Markov, a transition to the next state depends only on the previous state. 
%, i.e.,
%\begin{equation}
%  P(S_t = s \vert S_1 = s_1, S_2 = s_2, \ldots, S_{t-1} = s_t) = P(S_t = s \vert %S_{t-1} = s_t) \, .
%\end{equation}
The goal of the agent is to find an optimal way of behaving, i.e., a policy $\pi$ that maximizes the discounted cumulative reward, or return,
\begin{equation} \label{eq:return}
    R_t = \sum_{k=0}^{T-t-1}\gamma^k r_{t+1+k},
\end{equation}
where $\gamma\in [0,1]$ is a constant called the discount factor. The value of the discount factor indicates how far in time we take the rewards into account: if it is close to $0$, then we have a ``myopic'' vision of the process, while if it is close to $1$ then also actions that happen farther in time are relevant in the computation of the return. 
\begin{comment}
This leads us to the definition of the value of a state $S$
\begin{equation}
    v_{\pi}(s) = \mathbb{E}_{\pi}[R_t \vert s_t=s],
\end{equation}
where $\mathbb{E}_{\pi}$ means the expected value provided that the agent follows the policy $\pi$ at each $t$. Roughly speaking, the value of a state measures how good it is to be in that state. So the agent's goal is to determine an optimal policy $\pi_*$ such that
\begin{equation}
    v_{\pi}(S) \leq v_{\pi_*}(S)\quad \forall S \in \mathcal{S},
\end{equation}
for every policy $\pi$.
\end{comment}

In practice it is very difficult to solve an MDP exactly. For example, very often the %transitions
transition probabilities are not known and determining them
%\todo{Esmond:  It is not clear what ``values'' are referring to here.}
is computationally too expensive even for small scale problems. Thus one often tries to find an approximate solution of the MDP that models the reinforcement learning problem. In particular, we are going to estimate the probabilities by using a deep neural network. In this case, the policy will be denoted by $\pi_{\theta}$, where $\theta$ denotes the parameters of deep the neural network. 

 We make use of a popular policy gradient approach called synchronous advantage actor-critic (A2C)~\cite{reinforce} to find an approximate solution to the graph partitioning problem. A2C combines the standard REINFORCE algorithm~\cite{reinforce}, in which updates are made in the direction $\nabla_{\theta}\log(\pi_{\theta}(a_t, s_t))R_t$,
%\todo{Esmond:  Is $G_t$ supposed to be there?  If so, it has not been defined.}
with $\theta$ the parameters of the approximator, with a baseline $(R_t - b_t(s_t))$, that helps to reduce the variance. So the resulting update takes the form $\nabla_{\theta}\log(\pi_{\theta}(a_t, s_t)(R_t - b_t(s_t))$. A baseline that is commonly used is the value function of a state $v(s_t)$ \cite{suttonrl2018,baselineValue}, a function of a state that, roughly speaking, measures how good it is to be in that state. the term $(R_t - v(s_t) )$ is called the advantage. In this model, policy (actor) and value (critic) learning are highly intertwined, resulting in a lower variance and better stability of the approximation. The loss function to be minimized is then
\begin{equation}
    L = -\sum_{t=0}^{T-1}\log \pi_{\theta}(a_t\vert s_t)(R_t - v(s_t)).
\end{equation}
%\todo{Esmond:  Is there an inconsistency in notation?  It is not clear what the difference is between $\pi_{\theta}(a_t,S_t)$ and $\pi(a_t|S_t)$.}

In our case, the agent will be modeled by a two-headed deep neural network, with parameter tensor $\theta$, that takes as input a state $s$ and returns a tensor of probabilities $\pi_{\theta}(\cdot\vert s)$ for the actions, and a scalar value $v_{\theta}(s)$ for the value function. Here we use $v_{\theta}(s)$ to approximate the true value function $v(s_t)$. 
%Then we can mention like "notice that in our neural network architecture, we construct $A_{\theta}(s)$ by putting another single layer network followed by a global mean pooling layer on the top of $\pi_{\theta}(\cdot)$ to formulate the final scalar output of $A_{\theta}(s)$. In this fashion, both actor and critic networks share most of the model parameters, which drastically saves computation" }. 
Hence, the loss function becomes
\begin{equation}\label{eq:lossFunction}
    L(\theta) = -\sum_{t=0}^{T-1}\log \pi_{\theta}(a_t\vert s_t)(R_t - v_{\theta}(s_t)) + \alpha \sum_{t=0}^{T-1}(R_t - v_{\theta}(s_t))^2,
\end{equation}
where we added the weighted ``critic loss'' $\alpha \sum_{t=1}^{T-1}(R_t - v_{\theta}(s_t))^2$ to stabilize the training, with $\alpha \in (0,1]$.

\begin{figure}
    \centering
    \includegraphics[width=.7\textwidth]{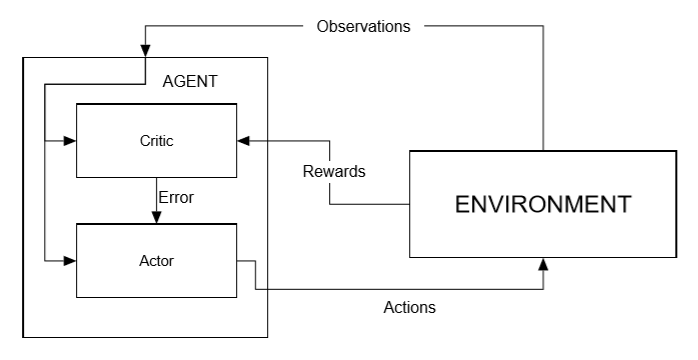}
    \caption{Illustration of the advantage actor-critic (A2C) reinforcement learning approach. Given a state $S_t$, the actor determines what action to take, generating a new state. Meanwhile, the critic branch of the network estimates the value ($v(S_t)$) of the state, which is used to compute the advantage ($R_t - v(S_t)$) of the future state $S_t$. This advantage is then used to reinforce the chosen action $a_t$. This figure is based on an illustration from~\cite{zai2020deep}.}
    \label{fig:a2c}
\end{figure}

% \todo[inline]{finish writing this section}
Figure \ref{fig:a2c} illustrates the A2C procedure. Given a state, the actor determines what action to take, which will generate a new state. Meanwhile, the critic branch of the network computes the value $v_{\theta}(s_t)$ of the state, which is used to compute the advantage ($R_t - v_{\theta}(s_t)$) of the future state $s_t$. This advantage is then used to reinforce the chosen action $a_t$. Algorithm \ref{alg:agent_update} illustrates the A2C weight update procedure.

\begin{algorithm}
	\textbf{Input:} model parameters $\theta$,\\
	\phantom{spacese} rewards $r_i$, log-probabilities $\log(\pi_i)$ and values $v_i$ for $i \in [1, t]$,\\
	\phantom{spacese} learning rate $\delta$, coefficient $\alpha$
	% \textbf{Output:} 
	\begin{algorithmic}[1]
	    \Procedure{update\_model\_parameters\_A2C}{$\theta$, $r$, $\log(\pi)$, $v$}
	        \State $R \gets \left[ r_1, \, r_2 + \gamma r_1, \, \ldots \, , r_t + \gamma r_{t-1} + \ldots + \gamma^{t-1} r_1 \right]$ \hfill \Comment{compute returns,~Eq. \ref{eq:return}}
            \State $R \gets \text{normalize}(R)$
            % \State $L_{\text{actor}} \gets - \text{logprobs} \cdot (R - V)$
            % \State $L_{\text{actor}} \gets - \log(\pi) \cdot (R - V)$
	        % \State $L_{\text{critic}} \gets (V - R)^2$
	        % \State $L \gets \sum{L_{\text{actor}}} + \alpha \sum{L_{\text{critic}}}$ 
	        % \State $\Delta \Theta \gets \text{backward}(L)$
	        \State \hfill \Comment{combine actor and critic loss, sum over steps taken,~Eq. \ref{eq:lossFunction}}
	        \State $L \gets -\sum_{1}^{t}{\log(\pi) (R - v) + \alpha \sum_1^t{(v - R)^2}}$
	        \State $\theta \gets \theta - \delta \cdot \partial L / \partial \theta$ \hfill \Comment{back-propagation and step with the optimizer}
	    \EndProcedure
	\end{algorithmic}
	\caption{Update model parameters $\theta$ using advantage actor critic (A2C).}
	\label{alg:agent_update}
\end{algorithm}

We set the coefficient $\alpha$ to $0.1$, in order to let the actor learn faster than the critic. The discount factor $\gamma$ is set to $0.9$, since we are interested in the long term return, as opposed to finding a strictly greedy approach. For the training we use the Adam optimizer \cite{adam2015} with learning rate $\delta$ set to $10^{-3}$.

% \subsection{Agent Neural Network\label{sec:network}}

\begin{algorithm}
	\textbf{Input:} graph $G(V, E)$, feature tensor $F$, $I_{\text{mask}}$ list of features with masks for nodes to exclude \\
	\textbf{Output:} actor (action (log-)probabilities),  critic (scalar value, only returned in training mode)
	\begin{algorithmic}[1]
		\Function{agent}{$G$, $F$, $I_{\text{mask}}$}
		    \State mask $ \gets $ where($\bigcup\limits_{i \in I_{\text{mask}}} (F(:, i) \neq 0))$ \label{line:create_mask}
		    %\For{$i \gets 1 \text{ to } \ell$}
		    %    \State $F \gets $ Tanh(GATConv($F$, $u$)) \hfill \Comment{convolutional layers common to actor and critic}
		    %\EndFor
		    \State $F \gets $ Tanh(SAGEConv($F$)) \hfill \Comment{convolutional layers common to actor and critic}
		    \State $F \gets $ Tanh(SAGEConv($F$))
		    \State actor $ \gets $ SAGEConv($F$) \hfill \Comment{actor branch}
		    % \State actor $ \gets $ SAGEConv(actor)
		    % \State actor $ \gets $ Linear(actor)
		    \State actor(mask) $ \gets -\infty$ \label{line:mask_infty} \hfill \Comment{exclude certain actions}
		    \State actor $ \gets $ log\_softmax(actor)
		    \If{not training}
		        \State \Return actor \hfill \Comment{critic is not required in eval mode}
		    \EndIf
		    % \State critic $ \gets F$ 
		    % \For{$i \gets 1 \text{ to } \ell_c$}
		      %  \State critic $ \gets $ Tanh(GATConv(critic, $u$)) \hfill \Comment{critic branch}
		    %\EndFor
		    \State critic $ \gets $ detach($F$) \hfill \Comment{critic branch} \label{line:detachF}
		    \State critic $ \gets $ Tanh(SAGEConv(critic))
		    \State critic $ \gets $ Linear(critic) \hfill \Comment{linear layer}
		    \State critic $ \gets $ Tanh(global\_mean\_pool(critic)) \hfill \Comment{reduce critic to a single scalar $\in (-1, 1)$} \label{line:global_pool_critic}
		    \State \Return actor, critic
	    \EndFunction
	\end{algorithmic}
	\caption{Forward evaluation of the agent's neural network.}
	\label{alg:agent_forward}
\end{algorithm}

Algorithm \ref{alg:agent_forward} shows the neural network used for all experiments, which relies on the SAGEConv graph convolutional layer, described in~\cite{hamilton2017inductive}. The SAGE layer implements the relation
\begin{equation}
    F'_i \gets F_i W_1 + (\text{mean}_{j \in \mathcal{N}(i)}F_j) W_2,
\end{equation}
where $F$ is the feature tensor, $W_1$ and $W_2$ are weight matrices and $\mathcal{N}(i)$ is the neighborhood of node $i$. The number of rows in $W_1$ and $W_2$ is determined by the number of features in $F$. In Algorithm \ref{alg:agent_forward}, the output dimensions are not explicitly mentioned. In the experiments, the number of output channels for the layers is set equal to the number of input channels. The network, as used with $5$ input features in Section \ref{sec:edge_separator} (finding an edge separator) has 182 tunable parameters. For computing a vertex separator, Section \ref{sec:vertex_separator}, $7$ input features are used and the network has $338$ parameters. The hyperbolic tangent is used for the nonlinear activation function. The network has two branches, the actor and the critic. The actor contains (log-)probabilities for the possible actions, while the critic predicts the value of the current state. Note that the critic is only required during training. The input to the critic branch of the network, $F$, is detached in Line \ref{line:detachF} so that the common layers are only updated by the actor loss. In Line \ref{line:global_pool_critic}, the critic is reduced to a single scalar value, which, due to the hyperbolic tangent activation function belongs to the interval $(-1,1)$.

The possible actions are each of the nodes in the graph. What each of these actions means will be discussed in the subsequent sections. However, depending on the context and the current state, certain nodes should not be chosen as they would not lead to valid actions. Therefore, some inputs to the log\_softmax layer are set to $- \infty$, so these nodes will never be selected as actions, see Line \ref{line:create_mask} and Line \ref{line:mask_infty}.

We implement Algorithm \ref{alg:agent_forward} using PyTorch geometric~\cite{Fey_Lenssen_2019}. To speed up the training process, we run the A2C training using multiple workers with a single shared model, which is called distributed A2C. The workers are created using the PyTorch multiprocessing API.

\section{Finding a Minimal Edge Separator\label{sec:edge_separator}}
Given a graph $G=(V,E)$, partitioned as $V = V_A \cup V_B$ with $V_A \cap V_B = \emptyset$, its cut is defined as
\begin{equation}
    \text{cut}(G) = \sum_{\substack{(v,w)\in E \\ v\in V_A,w\in V_B}}1 \, , \label{eq:cut}
\end{equation}
which is simply the number of edges between partitions $V_A$ and $V_B$.
(In the remainder of the paper, when we say $G=(V,E)$ is partitioned into $V_A$ and $V_B$, it is assumed that $V = V_A \cup V_B$ and $V_A \cap V_B = \emptyset$ even though we will not state it explicitly.)
Our goal is to minimize the cut while keeping the two partitions balanced. A popular objective is to minimize the normalized cut, which is defined as
\begin{equation}
    \text{NC}(G) = \text{cut}(G)\left(\frac{1}{\text{vol}(V_A)}+\frac{1}{\text{vol}(V_B)}\right) \, , \label{eq:normalized_cut}
\end{equation}
with the volume of a partition defined as
\begin{equation}
  \text{vol}(V_A) = \sum_{v \in V_A}\text{deg}(v) \, . \label{eq:volume}
\end{equation}
For a pictorial example of the cut and the normalized cut see Figure \ref{fig:minimumCut}.

\begin{algorithm}
	\textbf{Input:} graph $G(V, E)$ \\
	\textbf{Output:} partitions $V_A$ and $V_B$, such that $V = V_A \cup V_B$
	\begin{algorithmic}[1]
		\Function{edge\_separator}{$G$}
            \If{$|V| < n_{\text{min}}$} \hfill \Comment{end the recursion}
                \State \Return metis\_partition($G$) or edge\_separator\_coarse($G$) \label{line:metis_coarse} \hfill \Comment{See Section \ref{sec:edge_coarse_partitioning}}
            \EndIf
    		\State $G^C, I^C \gets $ coarsen($G$) \hfill \Comment{get coarse graph and interpolation info} \label{line:coarsening}
    		\State $V^C_A, V^C_B \gets $ metis\_partition($G^C$) if training else edge\_separator($G^C$) \hfill \Comment{recursion} \label{line:recursion}
    		%\If{training}
    		%    \State $V^C_A, V^C_B \gets $ metis\_partition($G^C$)
    		%\Else
    		%    \State $V^C_A, V^C_B \gets $ edge\_separator($G^C$) \hfill \Comment{recursive call using coarse graph}
    		%\EndIf
    		\State $V_A, V_B \gets V^C_A(I^C), V^C_B(I^C)$ \hfill \Comment{interpolate $V_A$ and $V_B$ from coarse to fine} \label{line:interpolation}
    		\State $V^0_A, V^0_B \gets V_A, V_B$ \hfill \Comment{keep a copy} \label{line:copyVAVB}
    		\State $V_C \gets \{ v \} \cup \{ w \}, \forall \, v, w : \exists \, e_{v,w} \text{ with } v \in V_A, w \in V_B$ \hfill \Comment{find nodes around the cut}
    		\State $G^{\text{sub}} \gets $ k\_hop\_subgraph($G$, $V_C$, $k_{\text{hops}}$) \hfill \Comment{subgraph with all nodes at most $k_{\text{hops}}$ from $V_C$} \label{line:khopsubgraph}
    		\State  \hfill \Comment{construct feature tensor}
	    	\State $F(v) \gets \left[ v \in V_A \, , v \in V_B \, , v \in \partial G^{\text{sub}} \, , \sfrac{\text{vol}(V_A)}{\text{vol}(V)} \, , \sfrac{\text{vol}(V_B)}{\text{vol}(V)} \right], \,  \forall \, v \in G^{\text{sub}}$ \label{line:Fconstruction}
    		\State $c \gets \text{cut}(G, V_A, V_B)$ \hfill \Comment{compute the cut size,~Eq. \ref{eq:cut}} \label{line:cut}
    		\State $C \gets \text{NC}(G, V_A, V_B)$ \hfill \Comment{compute normalized cut,~Eq. \ref{eq:normalized_cut}}
    		% \State $A, R \gets [\,], [\,]$ \hfill \Comment{actions, cumulative rewards}
    		\For{$t \gets 1 $ to $c$} \label{line:begin_episode}
	    	    \If{training}
	    	        \State policy$_t$, value$_t$ $ \gets $ agent($G^{\text{sub}}, F$, $2$) \hfill \Comment{forward evaluation of the agent, see~Algorithm \ref{alg:agent_forward}}
	    	        \State $a_t \gets $ categorical\_sample\_logits(policy$_t$) \hfill \Comment{pick action} \label{line:categorical_sampling}
	    	    \Else
	    	        \State policy$_t$ $ \gets $ agent($G^{\text{sub}}, F$, $2$) \hfill \Comment{forward evaluation of the agent, see~Algorithm \ref{alg:agent_forward}} \label{line:fwd_agent_edge}
	    	        \State $a_t \gets $ argmax(policy$_t$) \hfill \Comment{pick action} \label{line:argmax_policy}
	    	    \EndIf
	    	    % \State $A \gets \left[A, \, a \right]$
	    	    \State $V_A, V_B \gets V_A \setminus a_t, V_B \cup a_t \text{ if } a_t \in V_A \text{ else } V_A \cup a_t, V_B \setminus a_t$ \hfill \Comment{move $a_t$ from $V_A$/$V_B$ to $V_B$/$V_A$} \label{line:take_action}
                \State $F(v) \gets \left[ v \in V_A \, , v \in V_B \, , v \in \partial G^{\text{sub}} \, , \sfrac{\text{vol}(V_A)}{\text{vol}(V)} \, , \sfrac{\text{vol}(V_B)}{\text{vol}(V)} \right], \,  \forall \, v \in G^{\text{sub}}$ \hfill \Comment{update features} \label{line:Fupdate}
                \State $C_{\text{old}},\, C \gets C,\, \text{NC}(G, V_A, V_B)$ \hfill \Comment{compute normalized cut,~Eq. \ref{eq:normalized_cut}}
	    	    \State $r_t \gets C_{\text{old}} - C$ \hfill \Comment{compute reward} \label{line:reward_nc}
	    	    %\If{training}
	    	    %    \State update\_model\_parameters\_A2C(agent, $r$, policy, value)
	    	    %\EndIf
	    	\EndFor \label{line:end_episode}
	    	\If{training}
	    	    \State update\_model\_parameters\_A2C(agent, $r$, policy, value) \label{line:update_model}
	    	\Else
	    	    \State $V_A, V_B \gets V^0_A, V^0_B$
	    	    \For{$t \gets 1$ to argmax($r$)} \label{line:begin_application}
	      	        % \State $a \gets A(i)$
	      	        \State $V_A, V_B \gets V_A \setminus a_t, V_B \cup a_t \text{ if } a_t \in V_A \text{ else } V_A \cup a_t, V_B \setminus a_t$
	    	    \EndFor \label{line:end_application}
	    	\EndIf
	    	\State \Return $V_A, V_B$
		\EndFunction
	\end{algorithmic}
	\caption{Computing an edge separator using deep reinforcement learning. This illustrates both the training and evaluation on a single graph.}
	\label{alg:edge_separator}
\end{algorithm}

In Algorithm \ref{alg:edge_separator}, we present an algorithm to find a partition $V = V_A \cup V_B$ for the graph $G(V,E)$ that approximately minimizes the normalized cut. The outline of the algorithm is as follows. If the graph $G$ is large enough, the algorithm constructs a coarse representation $G^C$ of the graph, and Algorithm \ref{alg:edge_separator} is applied recursively on this coarser graph $G^C$. To stop the recursion, when the graph is small enough, i.e., when $|V| \leq n_{\text{min}}$, the graph is partitioned directly using for instance the METIS graph partitioner (Line \ref{line:metis_coarse}), or a separately trained reinforcement learning-based graph partitioning algorithm, see Section \ref{sec:edge_coarse_partitioning}. The result of the recursive call is a partitioning $V^C = V^C_A \cup V^C_B$ of the coarser graph, which is then interpolated back (Line \ref{line:interpolation}) to the finer graph $G$. The resulting partitioning, which should already be relatively good, is then refined using a deep reinforcement learning approach (Line \ref{line:begin_episode} - Line \ref{line:end_episode}). This multilevel approach is sketched in Figure \ref{fig:coarseningRefining}. However, to speed up the refinement process, a subgraph $G^{\text{sub}}$ is constructed (Line \ref{line:khopsubgraph}) with the nodes in $G$ that are within a small number of hops from the cut. The cut is then only refined within this subgraph.
%\todo{Esmond:  Is $[v]$ in line 9 of Algorithm 3 (and elsewhere) a set containing $v$?  Also, what is the difference between $G_{sub}$ and $G^{sub}$?}

For the coarsening step, we use the graclus~\cite{auer2012gpu} graph clustering code, which is based on heavy edge matching and groups nodes in clusters of size $2$, with a small number of unmatched nodes leading to clusters of size $1$. These clusters define the nodes for the coarse graph $G^C$. The coarsening rate is typically slightly less than $2$, leading to $\sim \log{|V|}$ recursion levels. For the interpolation of the coarse partitioning back to the finer graph, nodes in $G$ which correspond to coarse nodes in $V_A^C$, or $V_B^C$, are all assigned to $V_A$, or $V_B$ respectively, in the fine graph (Line \ref{line:interpolation}). In Algorithm \ref{alg:edge_separator}, $I^C$ denotes the mapping from coarse to fine nodes. Note that the first step in the graclus clustering algorithm is a random permutation of the nodes, which means that the graph coarsening phase is not deterministic. Since the partition quality depends also on the graph coarsening, for each graph the algorithm is repeated 3 times and the best partitioned graph is kept.

The procedure in Algorithm \ref{alg:edge_separator} handles both the training and the evaluation for a single graph. During the training Algorithm \ref{alg:edge_separator} is called for each graph in the training dataset, and this is repeated for multiple epochs. 

\subsection{Refinement of the Cut using Deep Reinforcement Learning}

Line \ref{line:begin_episode} to Line \ref{line:end_episode} of Algorithm \ref{alg:edge_separator} show a single episode of the deep reinforcement learning algorithm to refine the interpolated partition. Let $c$ denote the size of the cut of the interpolated partitioning, computed in Line \ref{line:cut}. Since it is assumed that the algorithm achieved a high quality partitioning on the coarser problem, we expect to only require $\mathcal{O}(c)$ steps to refine the cut on the finer graph in order to overcome imperfections introduced by the interpolation procedure. Therefore we set the episode length to $c$ (Line \ref{line:begin_episode}). From the experiments we observe that taking more than $c$ steps typically does not further improve the partitioning.
%\textbf{ \color{red} Zee: requiring only O(c) steps is a very inspiring statement. Is there any reference we can cite here? Or we may add one sentence like "the efficiency of using c episode length has been testified in the experiments listed later"}
In every step of the episode one node from $G^{\text{sub}}$ is selected and used to perform an action. During training, this node is selected by sampling (Line \ref{line:categorical_sampling}) from the agent's policy. The policy contains log-probabilities for each of the nodes in $G^{\text{sub}}$. The policy corresponds to the actor output from the agent's neural network (see Algorithm \ref{alg:agent_forward}) applied to the graph $G^{\text{sub}}$ and the corresponding node feature tensor $F$. The feature tensor is discussed in more detail below.  During evaluation, the node with highest probability is selected (Line \ref{line:argmax_policy}). When a node $a_t$ is selected, an action is taken (Line \ref{line:take_action}). The action at step $t$ is also denoted as $a_t$, and we use $a$ to refer to the vector with all actions from step $1$ up to the current step $t$. If the selected node $a_t$ is in $V_A$, then it is moved to $V_B$, alternatively, if $a_t$ was in $V_B$, it is moved to $V_A$. At this point the normalized cut is computed for the new state, and the reward $r_{t+1}$ at step $t$ is defined as the difference between the previous and the new normalized cut (Line \ref{line:reward_nc}). Note that if $a_t$ is moved from $V_A$ to $V_B$, the volumes $\text{vol}(V_A)$ and $\text{vol}(V_B)$ can simply be updated by subtracting and adding $\text{deg}(a_t)$ respectively. Likewise, the cut $c$ can be updated cheaply by only considering the neighbors of $a_t$.

\paragraph{The Feature Tensor}
At Line \ref{line:Fconstruction} the feature tensor $F$ is constructed, with $5$ features per node in the graph $G^{\text{sub}}$. The first two features are a one-hot encoding of the partition the node belongs to, with $\left[1, \, 0\right]$ referring to $V_A$ and $[0, \, 1]$ to $V_B$. Let $\partial G^{\text{sub}}$ denote the boundary of the subgraph $G^{\text{sub}}$ around the cut, i.e., the nodes in $G^{\text{sub}}$ which are connected to nodes in $G \setminus G^{\text{sub}}$. These nodes have edges which are not part of $G^{\text{sub}}$ and are hence not seen by the agent. Therefore, these nodes should not be selected as actions. Thus, the next feature denotes whether ($1$) or not ($0$) a node is in $\partial G^{\text{sub}}$. Nodes with this feature set to $1$ will never be selected, since in the agent's neural network their input to the softmax %\todo{Esmond: Is it clear what ``softmax'' is?} 
layer is set to minus infinity, see Line \ref{line:create_mask}, \ref{line:mask_infty} in Algorithm \ref{alg:agent_forward}. In our case, the softmax layer is defined as a layer $\sigma$ that takes as input a $(m,1)$-tensor $z$ and returns
\begin{equation}
    \sigma(z)_i=\frac{\exp(z_i)}{\sum_{j=1}^m \exp(z_j)},
\end{equation}
where $m$ is the number of nodes of the input graph $G^{\text{sub}}$.
The final two features are the normalized volumes of the two partitions: $\text{vol}(V_A)/\text{vol}(V)$ and $\text{vol}(V_B)/\text{vol}(V)$. These volumes play an important role in the reward function, and they cannot be determined from $G^{\text{sub}}$ and the other features alone. The feature tensor is updated in every step of the episode, after an action is taken, on Line \ref{line:Fupdate}. Table \ref{table: featureTensors} summarizes the different features used for each deep reinforcement learning model.

\begin{table}
\begin{center}
\begin{tabular}{cccccccc}

Algorithm & \multicolumn{7}{c}{Feature tensor} \\

\hline
 & \multicolumn{3}{c}{Partition info} & \multicolumn{2}{c}{Binary mask} & \multicolumn{2}{c}{Imbalance info}\\
 \cmidrule(lr){2-4}\cmidrule(lr){5-6}\cmidrule(lr){7-8}
 %& \multicolumn{3}{c}{Partition info} & \multicolumn{2}{c}{Binary mask} & \multicolumn{2}{c}{Imbalance info}\\
Edge-cut refining & $ v \in V_A$ & $ v \in V_B$ & & $v \in \partial G^{\text{sub}} $ &  &  $\frac{\text{vol}(V_A)}{\text{vol}(V)}$ & $\frac{\text{vol}(V_B)}{\text{vol}(V)}$ \\
Edge-cut coarsest graph & $ v\in V_A$ & $v\in V_B$ & & & & & \\
Vertex separator & $ v \in V_A$ & $ v \in V_B$ & $ v \in V_S$ & $ v \in \partial G^{\text{sub}}$ & $ v \in V_S^{\min}$ &  $ \sfrac{|V_A|}{|V|}$ & $ \sfrac{|V_B|}{|V|}$\\ \hline
\end{tabular}
\end{center}
\caption{Feature tensors for each considered model. On the left  we list the studied models: edge-cut refining, coarsest graph partitioning and vertex separator. On the right, each feature is positioned in a specific column according if it is related to the partition information, it serves as a mask or it is an imbalance information. Binary mask features exclude the node from being picked during the DRL process.} \label{table: featureTensors}
\end{table}

\paragraph{}
In Algorithm \ref{alg:edge_separator}, $r$ is used to denote the vector of all rewards from the first steps up to the current step $t$. Similarly, policy$_t$ refers to the log-probabilities at step $t$ and policy contains all these log-probabilities stacked together. In Algorithm \ref{alg:edge_separator}, to simplify the notation, the network parameters are updated only once, after the entire episode is finished. The update of the model parameters is done in Line \ref{line:update_model}, with a call to the A2C update procedure Algorithm \ref{alg:agent_update}. However, in practice, the network parameters $\theta$ are updated after a fixed number of steps in the episode, as well as at the end of the episode.

In evaluation mode, only the actions that actually contribute to the peak cumulative reward are applied to the graph. Therefore, a copy is made of the initial partitioning $V_A$, $V_B$ before the start of the episode, see Line \ref{line:copyVAVB}. Line \ref{line:begin_application} to Line \ref{line:end_application} apply the actions from the episode that lead to the peak cumulative reward obtained during the episode.

\subsection{Partitioning the Coarsest Graph\label{sec:edge_coarse_partitioning}}
\begin{figure}
    \centering
    \includegraphics[scale=0.8]{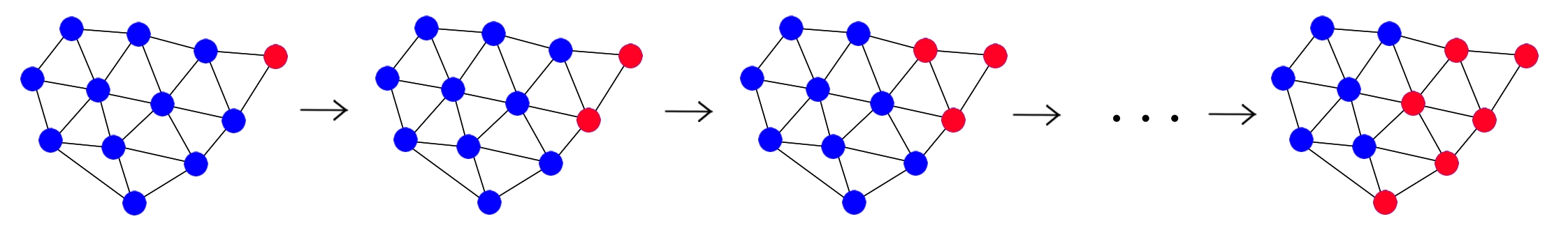}
    \caption{Representation of the algorithm to partition the coarsest graph. At the first step all nodes are in one partition (blue) except for one node with minimum degree (the red one). Then, at each step the agent picks a blue node and turns it into a red one, until there are exactly the same number of blue and red vertices.}
    \label{fig:partitioningCoarsestGraph}
\end{figure}

In order to partition the coarsest graphs (see Line \ref{line:metis_coarse} in Algorithm \ref{alg:edge_separator}) we train a separate actor-critic reinforcement learning agent. Given a graph $G$ with $n$ vertices, we assign all of its nodes to partition $B$, except for one, one of the nodes with smallest degree, which is assigned to partition $A$. By choosing the node with smallest degree the initial normalized cut is as small as possible. In this case, each node $v$ in $B$ has feature $F(v)=[1,0]$, while each node $v$ in $A$ has feature $F(v)=[0,1]$, see Table \ref{table: featureTensors}. During the training, the agent picks a node in partition $B$, according to the output probabilities of the deep neural network, and moves it to partition $A$. This corresponds to changing the feature $F(v)$ of the chosen node from $[1,0]$ to $[0,1]$. The process is repeated until the two partitions reach the same cardinality: $|A| = |B|$ if $n$ is even or $|A|=|B|+1$ if $n$ is odd. During evaluation we allow an imbalance of $\iota = 1\%$ of $n$ between the two partitions. Algorithm \ref{alg:edge_separator_coarsest} shows the training and the evaluation process explained above. Figure \ref{fig:partitioningCoarsestGraph} illustrates the above algorithm on a toy example.

\begin{algorithm}
	\textbf{Input:} graph $G(V, E)$, with $\vert V\rvert=n$. Imbalance factor $\iota$. \\
	\textbf{Output:} partitions $V_A$ and $V_B$, such that $V = V_A \cup V_B$
	\begin{algorithmic}[1]
		\Function{edge\_separator\_coarse}{$G$}
    		\State $V^0_A, V^0_B \gets V_A, V_B$ \hfill \Comment{keep a copy} \label{line:copyVAVB_coa}
    		\State $C \gets \text{NC}(G, V_A, V_B)$ \hfill \Comment{compute normalized cut,~Eq. \ref{eq:normalized_cut}}
    		\For{$t \gets 1 $ to $\sfrac{n}{2}-1$} \label{line:begin_episode_coa}
	    	    \If{training}
	    	        \State policy$_t$, value$_t$ $ \gets $ agent($G, F$, $2$) \hfill \Comment{forward evaluation of the agent, see~Algorithm \ref{alg:agent_forward_coa}}
	    	        \State $a_t \gets $ categorical\_sample\_logits(policy$_t$) \hfill \Comment{pick action} \label{line:categorical_sampling_coa}
	    	    \Else
	    	        \State policy$_t$ $ \gets $ agent($G, F$, $2$) \hfill \Comment{forward evaluation of the agent, see~Algorithm \ref{alg:agent_forward_coa}}
	    	        \State $a_t \gets $ argmax(policy$_t$) \hfill \Comment{pick action} \label{line:argmax_policy_coa}
	    	    \EndIf
	    	    % \State $A \gets \left[A, \, a \right]$
	    	    \State $V_A, V_B \gets V_B \setminus a_t, V_A \cup a_t $ \hfill \Comment{move $a_t$ from $V_A$/$V_B$} \label{line:take_action_coa}
                \State $F(a_t) \gets [0,1]$ \hfill \Comment{update features} \label{line:Fupdate_coa}
                \State $C_{\text{old}},\, C \gets C,\, \text{NC}(G, V_A, V_B)$ \hfill \Comment{compute normalized cut,~Eq. \ref{eq:normalized_cut}}
	    	    \State $r_t \gets C_{\text{old}} - C$ \hfill \Comment{compute reward}
	    	    %\If{training}
	    	    %    \State update\_model\_parameters\_A2C(agent, $r$, policy, value)
	    	    %\EndIf
	    	\EndFor \label{line:end_episode_coa}
	    	\If{training}
	    	    \State update\_model\_parameters\_A2C(agent, $r$, policy, value) \label{line:update_model_coa}
	    	\Else
	    	    \State $V_A, V_B \gets V^0_A, V^0_B$
	    	    %\State imbalance $ \gets \sfrac{\iota n}{100}$
	    	    \For{$t \gets 1$ to $n/2 - 1 + \iota n / 100$} \label{line:begin_application_coa}
	      	        % \State $a \gets A(i)$
	      	        \State $V_A, V_B \gets V_B \setminus a_t, V_A \cup a_t $
	    	    \EndFor \label{line:end_application_coa}
	    	\EndIf
	    	\State \Return $V_A, V_B$
		\EndFunction
	\end{algorithmic}
	\caption{Computing an edge separator using deep reinforcement learning on the coarsest graph. This illustrates both the training and evaluation on a single graph.}
	\label{alg:edge_separator_coarsest}
\end{algorithm}

The deep neural network used for this task is slightly different than the one (Algorithm \ref{alg:agent_forward}) for the refinement phase. An important ingredient is the attention mechanism on graphs, which is implemented by graph attention (GAT)~\cite{attention2018} layers. The neural network has the actor and the critic branches after $4$ common GAT layers. For the actor, a few dense layers are used, while for the critic branch we use a global pooling layer~\cite{globalattention2017}, which allows to get a scalar value. In this case, the nodes that have already been picked, i.e., the nodes with feature $[0,1]$, are masked. All convolutional layers have $10$ units, while all linear layers have $5$ units. The global pooling layer includes a dense neural network with $2$ linear layers, with $5$ and $1$ units respectively, and Tanh as activation function in between. Algorithm \ref{alg:agent_forward_coa} shows the structure of the neural network.

\begin{algorithm}
	\textbf{Input:} graph $G(V, E)$, feature tensor $F$, $I_{\text{mask}}$ list of features with masks for nodes to exclude \\
	\textbf{Output:} actor (action (log-)probabilities),  critic (scalar value, only returned in training mode)
	\begin{algorithmic}[1]
		\Function{agent}{$G$, $F$, $I_{\text{mask}}$}
		    \State mask $ \gets $ where($\bigcup\limits_{i \in I_{\text{mask}}} (F(:, i) \neq 0))$ \label{line:create_mask_coa}
		    \State $F \gets $ Tanh(GATConv($F$)) \hfill
		    \Comment{convolutional layers common to actor and critic}
		    \State $F \gets $ Tanh(GATConv($F$)) \hfill	
		    \State $F \gets $ Tanh(GATConv($F$)) \hfill		    
		    \State $F \gets $ Tanh(GATConv($F$)) \hfill		    
            \State $F\gets $ Tanh(Linear($F$)) \hfill		    
		    \Comment{dense layers common to actor and critic}
		    \State $F\gets $ Tanh(Linear($F$)) \hfill

		    \State actor $ \gets $ Tanh(Linear($F$)) \hfill \Comment{actor branch}
		    \State actor $ \gets $ Tanh(Linear(actor)) \hfill

		    \State actor(mask) $ \gets -\infty$ \label{line:mask_infty_coa} \hfill \Comment{exclude certain actions}
		    \State actor $ \gets $ log\_softmax(actor)
		    \If{not training}
		        \State \Return actor \hfill \Comment{critic is not required in eval mode}
		    \EndIf

		    \State critic $ \gets $ GlobalAttention($F$) \hfill \Comment{critic branch}
		    \State critic $ \gets $ Tanh(Linear(critic)) \hfill
		    \State critic $ \gets $ Linear(critic) \hfill %\Comment{linear layer} \label{line:global_pool_critic_coa}
		    \State \Return actor, critic
	    \EndFunction
	\end{algorithmic}
	\caption{Forward evaluation of the agent's neural network on the coarsest graph.}
	\label{alg:agent_forward_coa}
\end{algorithm}

\subsection{Algorithm Complexity\label{sec:edge_complexity}}
For a graph with $n = k^2$ nodes resulting from the spatial discretization of a regular, square, $k \times k$ two-dimensional problem, the size of the cut will be $c = \mathcal{O}(n^{\sfrac{1}{2}})$. Likewise, for a three-dimensional problem, the cut will be $c = \mathcal{O}(n^{\sfrac{2}{3}})$, i.e., a plane through the domain. In Algorithm \ref{alg:edge_separator}, the evaluation of the neural network (Line \ref{line:fwd_agent_edge}) on $G^{\text{sub}}$ takes $\mathcal{O}(c)$ computations, since $G^{\text{sub}}$ has $\mathcal{O}(c)$ nodes and the network has a fixed, small number of convolutional layers. The refinement procedure takes at most $\mathcal{O}(c)$ steps, so the total cost for the refinement at the finest level is $\mathcal{O}(c^2)$ or $\mathcal{O}(n)$ in 2D and $\mathcal{O}(n^{\sfrac{4}{3}})$ in 3D. Since the coarsening rate is close to $2$, the total cost for Algorithm \ref{alg:edge_separator} is 
\begin{equation}
    \text{cost}_{2D} = \sum_{i=1}^{\log n}{\frac{n}{2^{i-1}}} = \mathcal{O}(n) \, , \quad
    \text{cost}_{3D} = \sum_{i=1}^{\log n}{\left(\frac{n}{2^{i-1}}\right)^{\sfrac{4}{3}}} = \mathcal{O}(n^{\frac{4}{3}}) \, ,
\end{equation}
for 2D and 3D respectively.

%\todo[inline]{We can only say something if we make assumptions on the size of the cut? Because we take a subgraph around the cut.}

\subsection{Experimental Evaluation\label{sec:edge_experiments}}
This section discusses experiments with the proposed DRL partitioning algorithm. We consider two types of graphs: triangulations and graphs from a variety of different applications, all from spatial discretizations. For the first class, training is performed on a set of Delaunay triangulations and the resulting algorithm is tested on other triangulations, constructed using finite element modeling. In the second case, we train on graphs from 2D and 3D discretizations from a matrix collection commonly used as benchmark.

% \subsubsection{Triangulations}
%\todo[inline]{Explain that we train on Delaunay triangulations, then test on Delaunay and some other triangulations from Finite element modeling}

\paragraph{Delaunay Triangulations}
% \todo[inline]{describe this, parameters etc...}

\begin{figure}
    \centering
    \begin{subfigure}[b]{0.49\textwidth}
      \centering
      \includegraphics[width=\textwidth]{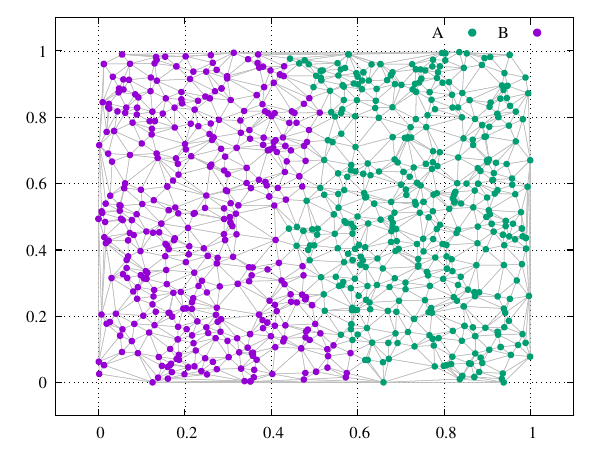}
    \caption{Partitioned Delaunay triangulation}
    \label{fig:delaunay_partition}
    \end{subfigure}
    \begin{subfigure}[b]{0.49\textwidth}
      \centering
      \includegraphics[width=\textwidth]{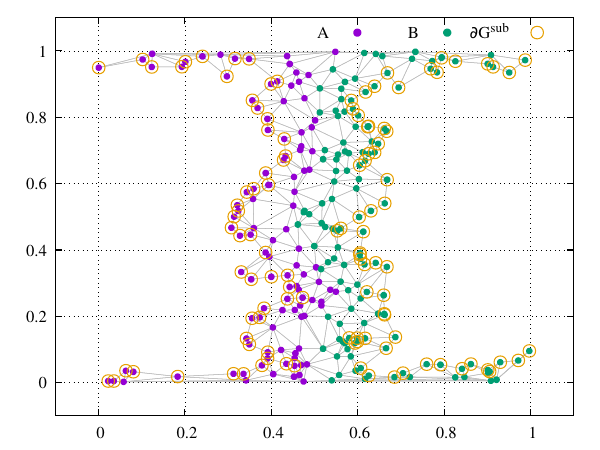}
      \caption{3-hop subgraph $G^{\text{sub}}$ around cut}
      \label{fig:Delaunay_partition_sub}
   \end{subfigure}
   \caption{Figure \ref{fig:delaunay_partition} shows a Delaunay triangulation from 750 random points in $[0, 1]^2$, partitioned using METIS. Call $V_A$ the partition including the purple nodes and $V_B$ the partition including the green nodes. Figure \ref{fig:Delaunay_partition_sub} illustrates the $3$-hop subgraph around the edge cut. The purple circular nodes have features $[1,0,0,\sfrac{\text{vol}(V_A)}{\text{vol}(V)} \, , \sfrac{\text{vol}(V_B)}{\text{vol}(V)}]$ while the green ones have features $[0,1,0,\sfrac{\text{vol}(V_A)}{\text{vol}(V)}]$. The nodes bounded by an orange circle belong to the boundary. More precisely, the purple orange-bounded nodes have features $[1,0,1,\sfrac{\text{vol}(V_A)}{\text{vol}(V)} \, , \sfrac{\text{vol}(V_B)}{\text{vol}(V)}]$, while the green orange-bounded nodes have features $[0,1,1,\sfrac{\text{vol}(V_A)}{\text{vol}(V)} \, ,\sfrac{\text{vol}(V_B)}{\text{vol}(V)}]$.}
   \label{fig:Delaunay_partition_with_sub}
\end{figure}

The graphs in this test correspond to planar Delaunay triangulations from points randomly generated in the unit square, see for example Figure \ref{fig:delaunay_partition}. The training dataset contains roughly $N^{\text{train}}$ graphs with $n^{\text{train}} \in [n^{\text{train}}_{\text{min}}, n^{\text{train}}_{\text{max}}]$ nodes. This set is constructed as follows. A random Delaunay graph with $n^{\text{train}}$ (sampled uniformly from $[n^{\text{train}}_{\text{min}}, n^{\text{train}}_{\text{max}}]$) nodes is generated and added to the dataset. Then this last graph is coarsened, and the coarsened graph is added to the dataset. As long as the coarser graph has more than $n^{\text{train}}_{\text{min}}$ nodes, this is repeated by coarsening the coarser graph again and adding the coarser version to the dataset as well. These operations are repeated until the dataset has $N^{\text{train}}$ elements. The graphs in the test set are simply randomly generated Delaunay triangulations, without the coarsenings. The parameter $n^{\text{train}}_{\text{min}}$ is also used to stop the recursive coarsening during the evaluation phase, see Line \ref{line:metis_coarse} in Algorithm \ref{alg:edge_separator}. 

We consider four different tests, with different parameters as detailed in Table \ref{tab:train_test_edge_params}.
\begin{table}
  \centering
  \begin{tabular}{ c c | c | c c c c }
    %& \multicolumn{2}{c|}{\ding{172}} & \multicolumn{2}{c|}{\ding{173}} & \multicolumn{2}{c}{\ding{174}} \\
    &  & Train & Test \ding{172} & Test \ding{173} & Test \ding{174} & Test \ding{175} \\
    \hline
    \multirow{4}{*}{$N$} & Delaunay & 10000 & 20 & 127 & 181 & 172 \\
     & GradedL & - & 16 & 26 & 6 & - \\
     & Hole3 & - & 11 & 20 & 13  & 5 \\
     & Hole6 & - & 11 & 20 & 13 & 5 \\
     \hline
    & $n_{\text{min}}$ & 100 & 100 & 5000 & 30000 & 60000 \\
    & $n_{\text{max}}$ & 5000 & 5000 & 30000 & 60000 & 90000 \\
    \hline
  \end{tabular}
  \caption{Parameters describing the datasets. The graphs in the Delaunay dataset are constructed from $n_{\text{min}} < n \leq n_{\text{max}}$ random points in $[0, 1]^2$. The GradedL, Hole3 and Hole6 dataset are finite element triangulations of 3 different geometries, for multiple levels of mesh refinement. The test sets are split in 4 separate ranges.}
  \label{tab:train_test_edge_params}
\end{table}

\begin{figure}
    \centering
    \begin{subfigure}[b]{0.49\textwidth}
      \centering
      \includegraphics[width=\textwidth]{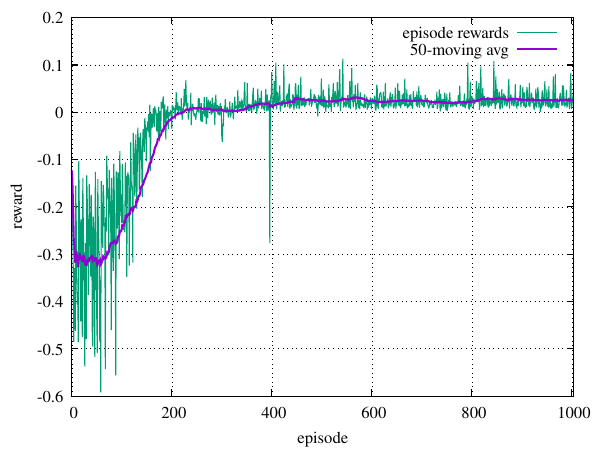}
    \caption{Episode rewards}
    \label{fig:edge_training_Delaunay_episode_rewards}
    \end{subfigure}
    \begin{subfigure}[b]{0.49\textwidth}
      \centering
      \includegraphics[width=\textwidth]{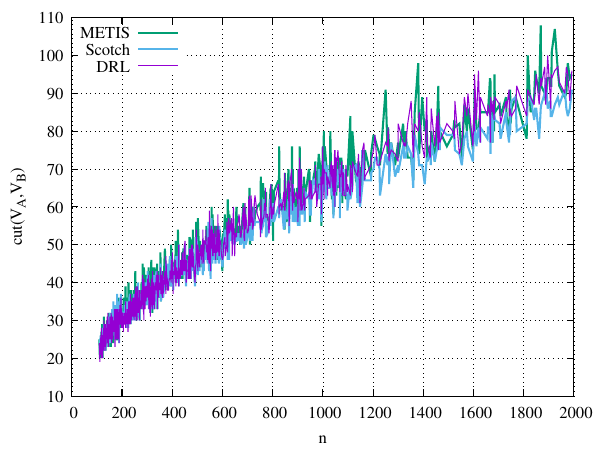}
      \caption{Edge cut}
      \label{fig:edge_training_Delaunay_rewards}
   \end{subfigure}
   \caption{Training on dataset with Delaunay graphs from random points in $[0, 1]^2$. Figure \ref{fig:edge_training_Delaunay_episode_rewards}: Total rewards of 1000 episodes. The final rewards are consistently positive. Figure \ref{fig:edge_training_Delaunay_rewards}: Cut sizes of all graphs in the training dataset, sorted based on number of nodes, as computed with the propsed DRL method, METIS and SCOTCH. For these 2D graphs, the cut size scales as $\sqrt{n}$.}
   \label{fig:Delaunay_rewards_cut}
\end{figure}

Figure \ref{fig:edge_training_Delaunay_episode_rewards}
shows the cumulative rewards for episodes on each of the graphs in the training dataset. Figure \ref{fig:edge_training_Delaunay_rewards} compares the cut sizes on the training graphs for the proposed method, METIS and SCOTCH. This also illustrates that for planar problems on a square domain, as expected, the cut size scales as $\sqrt{n}$.

\begin{figure}
    \centering
    \begin{subfigure}[b]{0.32\textwidth}
      \centering
      \includegraphics[width=1.05\textwidth]{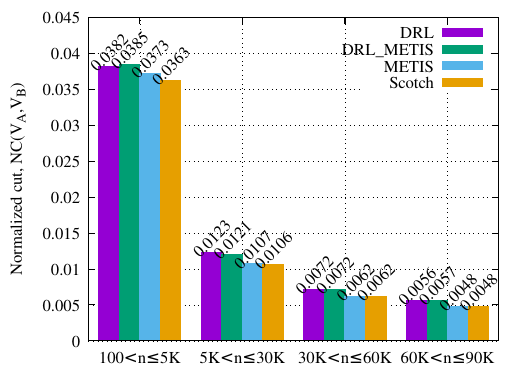}
    \caption{Normalized cut}
    \label{fig:normalized_cuts_Delaunay}
    \end{subfigure}
    \begin{subfigure}[b]{0.32\textwidth}
      \centering
      \includegraphics[width=1.05\textwidth]{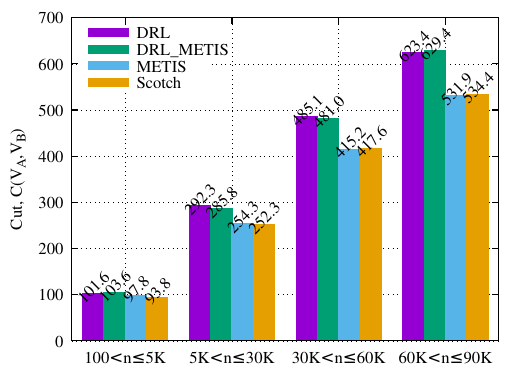}
      \caption{Edge cut}
      \label{fig:cuts_Delaunay}
   \end{subfigure}
   \begin{subfigure}[b]{0.32\textwidth}
      \centering
      \includegraphics[width=1.05\textwidth]{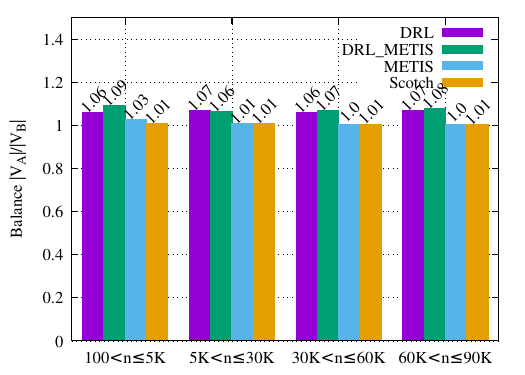}
      \caption{Partition balance}
      \label{fig:balance_Delaunay}
   \end{subfigure}
   \caption{Evaluation of the partitioning algorithms on the Delaunay testing sets. DRL and DRL\_METIS refer to Algorithm \ref{alg:edge_separator}. DRL\_METIS uses METIS on the coarses level, while DRL uses reinforcement learning partitioning on the coarsest level as well.}
   \label{fig:Delaunay_nc_c_balance}
\end{figure}

Figure \ref{fig:Delaunay_nc_c_balance} compares the normalized cut, the cut and the balance on the Delaunay testing dataset obtained by our model, METIS and SCOTCH. The balance is measured by $\max\left\{\frac{\text{vol}(A)}{\text{vol}(B)},\frac{\text{vol}(B)}{\text{vol}(A)}\right\}$. We see that, in every range of nodes, the normalized cut is very close to the one produced by METIS and SCOTCH and the partitions are balanced as well. The edge cut results are slightly higher for graphs having from $60,\!000$ to $90,\!000$ nodes.

\begin{figure}
    \centering
    \begin{subfigure}[b]{0.32\textwidth}
      \centering
      \includegraphics[width=1.05\textwidth]{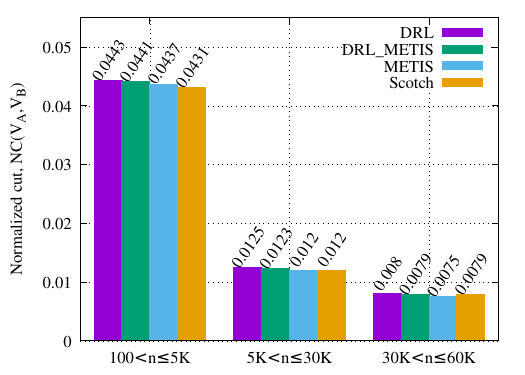}
    \caption{Normalized cut}
    \label{fig:normalized_cuts_GradedL}
    \end{subfigure}
    \begin{subfigure}[b]{0.32\textwidth}
      \centering
      \includegraphics[width=1.05\textwidth]{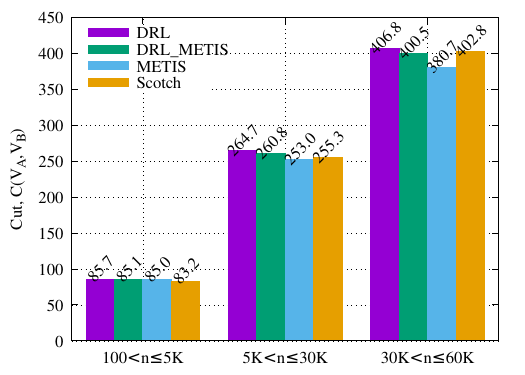}
      \caption{Edge cut}
      \label{fig:cuts_GradedL}
   \end{subfigure}
   \begin{subfigure}[b]{0.32\textwidth}
      \centering
      \includegraphics[width=1.05\textwidth]{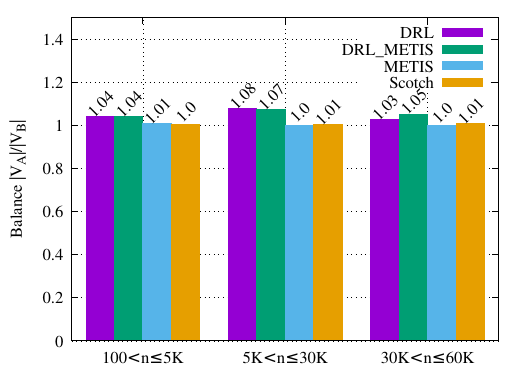}
      \caption{Partition balance}
      \label{fig:balance_GradedL}
   \end{subfigure}
   \caption{Evaluation of the partitioners on the GradedL triangulation dataset. Training was performed on Delaunay graphs with $100 < n \leq 5000$ nodes. }
   \label{fig:GradedL_nc_c_balance}
\end{figure}

\begin{figure}
    \centering
    \begin{subfigure}[b]{0.32\textwidth}
      \centering
      \includegraphics[width=1.05\textwidth]{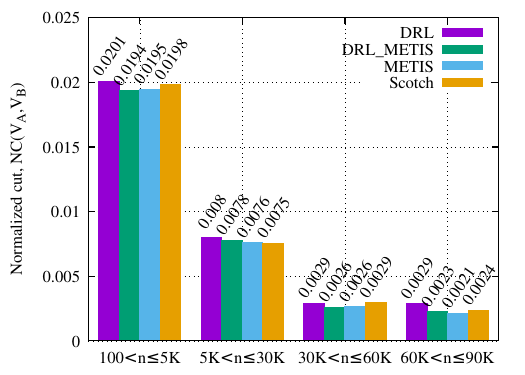}
    \caption{Normalized cut}
    \label{fig:normalized_cuts_Hole3}
    \end{subfigure}
    \begin{subfigure}[b]{0.32\textwidth}
      \centering
      \includegraphics[width=1.05\textwidth]{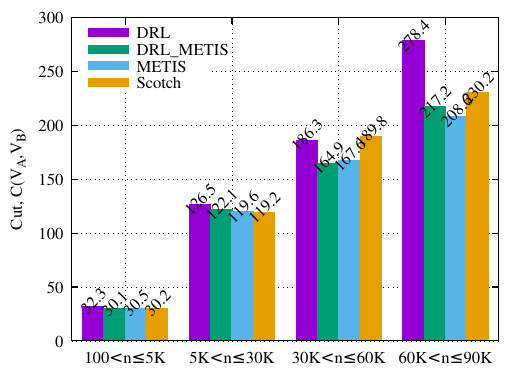}
      \caption{Edge cut}
      \label{fig:cuts_Hole3}
   \end{subfigure}
   \begin{subfigure}[b]{0.32\textwidth}
      \centering
      \includegraphics[width=1.05\textwidth]{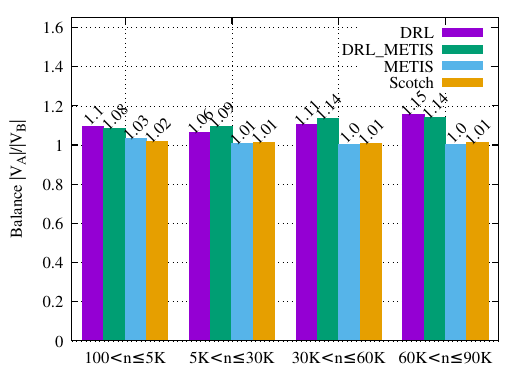}
      \caption{Partition balance}
      \label{fig:balance_Hole3}
   \end{subfigure}
   \caption{Evaluation of the partitioners on the Hole3 (see Figure \ref{fig:Hole3_partition})  triangulation dataset. Training was performed on Delaunay graphs with $100 < n \leq 5000$ nodes. }
   \label{fig:Hole3_nc_c_balance}
\end{figure}

\begin{figure}
    \centering
    \begin{subfigure}[b]{0.32\textwidth}
      \centering
      \includegraphics[width=1.05\textwidth]{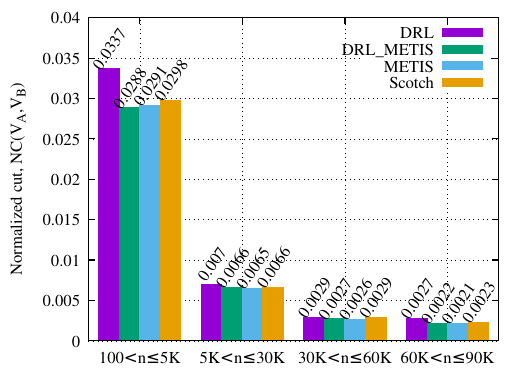}
    \caption{Normalized cut}
    \label{fig:normalized_cuts_Hole6}
    \end{subfigure}
    \begin{subfigure}[b]{0.32\textwidth}
      \centering
      \includegraphics[width=1.05\textwidth]{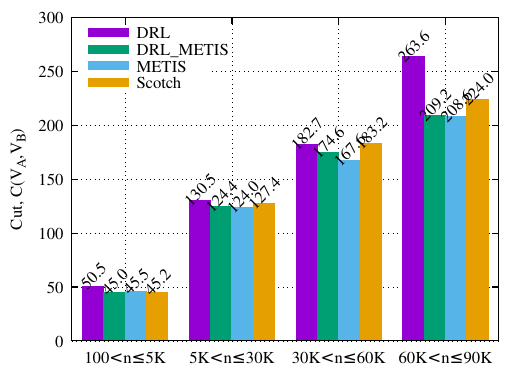}
      \caption{Edge cut}
      \label{fig:cuts_Hole6}
   \end{subfigure}
   \begin{subfigure}[b]{0.32\textwidth}
      \centering
      \includegraphics[width=1.05\textwidth]{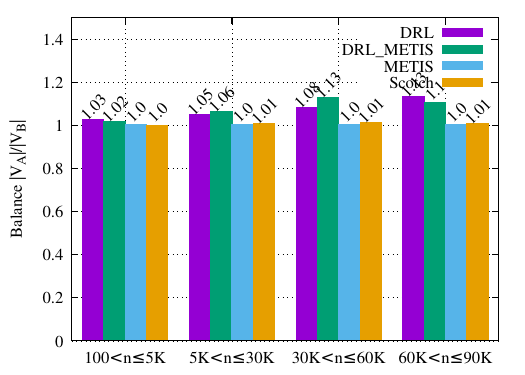}
      \caption{Partition balance}
      \label{fig:balance_Hole6}
   \end{subfigure}
   \caption{Evaluation of the partitioners on the Hole6 (see Figure \ref{fig:Hole6_partition}) triangulation dataset. Training was performed on Delaunay graphs with $100 < n \leq 5000$ nodes.}
   \label{fig:Hole6_nc_c_balance}
\end{figure}

Figure \ref{fig:Delaunay_partition_with_sub} illustrates the partitioning of one example of a Delaunay graph. Figure \ref{fig:Delaunay_partition_sub} shows the sub-graph $G^{\text{sub}}$, consisting of all nodes at distance at most $3$ hops from the edge cut.

\begin{figure}
    \centering
    \begin{subfigure}[b]{0.32\textwidth}
      \centering
      \includegraphics[width=\textwidth]{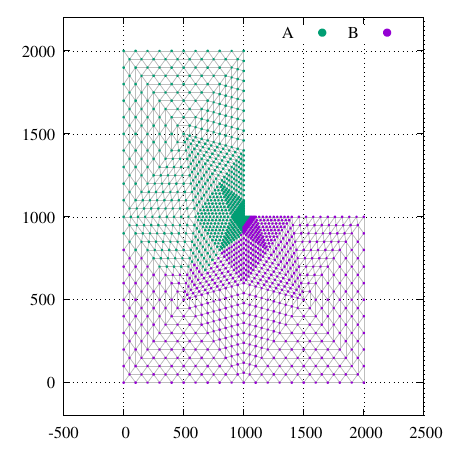}
      \caption{GradedL}
      \label{fig:GradedL_partition}
    \end{subfigure}
    \begin{subfigure}[b]{0.32\textwidth}
      \centering
      \includegraphics[width=\textwidth]{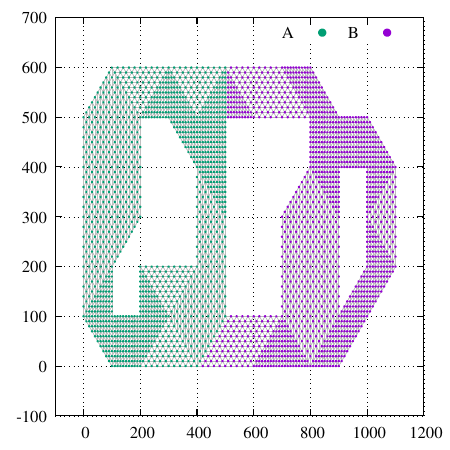}
      \caption{Hole3}
      \label{fig:Hole3_partition}
   \end{subfigure}
   \begin{subfigure}[b]{0.32\textwidth}
      \centering
      \includegraphics[width=\textwidth]{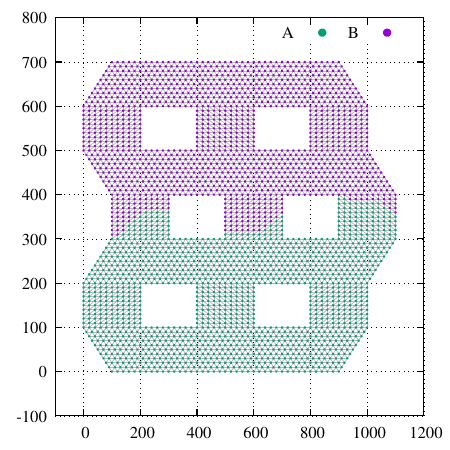}
      \caption{Hole6}
      \label{fig:Hole6_partition}
   \end{subfigure}
   \caption{Illustrations for the different finite element triangulations, Figure \ref{fig:GradedL_partition} GradedL after 15 refinements, Figure \ref{fig:Hole3_partition} Hole3 after 10 refinements, and Figure \ref{fig:Hole6_partition} Hole6 after 10 refinements.}
   \label{fig:GradedL_Hole_partition}
\end{figure}

% \todo[inline]{check in code the length of an episode}

\paragraph{Finite Element Triangulations}
% \todo[inline]{show the method generalizes}
Figure \ref{fig:GradedL_Hole_partition} illustrates the finite element triangulations we use for testing. We consider three different meshes -- GradedL, Hole3 and Hole6 
%from the MFEM library \cite{anderson2019mfem} 
-- each with multiple levels of refinement. Figures \ref{fig:GradedL_nc_c_balance}, \ref{fig:Hole3_nc_c_balance} and \ref{fig:Hole6_nc_c_balance} show the normalized cut, the cut size and the balance for these datasets, and the comparison with the ones obtained with METIS and SCOTCH. Notice that partitioning quality is close to that of METIS or SCOTCH. Recall that these graphs have a very different sparsity pattern with respect to the ones in the training dataset (Delaunay graphs), so this shows that the agent is able to generalize well on unseen planar triangulations.

\paragraph{Sparse Matrices from the SuiteSparse Collection}
%\todo[inline]{show some results}
We also tested our model on a number of 2D and 3D discretizations from the SuiteSparse Matrix Collection \cite{suitesparse}. We trained a separate agent, with a deep neural network having the same structure as the one used for the Delaunay triangulations. The training dataset is built in the same fashion as we did for the training on Delaunay triangulations. More precisely, we pick a (fully connected) 2D/3D discretization having number of vertices in $(100, 5000]$ and we add it to the dataset. Then we coarsen it until the coarsest graph has less than $100$ nodes and we add all the coarser graphs to the dataset. These operations are repeated until the dataset has $10,\!000$ elements. It is important to stress that training a separate agent for these kind of graphs is necessary, since they have a great variety of sparsity patterns. The agent trained on Delaunay triangulations may fail to generalize correctly, in particular in providing balanced partitions. 
 
\begin{figure}
    \centering
    \begin{subfigure}[b]{0.32\textwidth}
      \centering
      \includegraphics[width=1.05\textwidth]{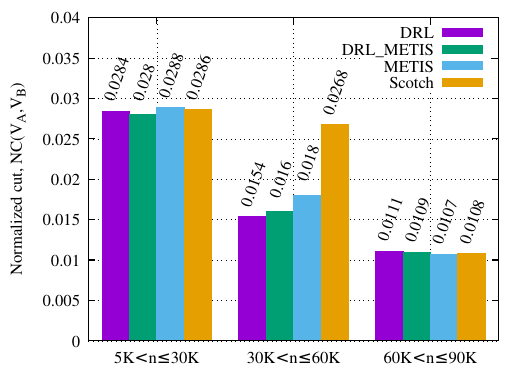}
    \caption{Normalized cut}
    \label{fig:normalized_cuts_SuiteSparse}
    \end{subfigure}
    \begin{subfigure}[b]{0.32\textwidth}
      \centering
      \includegraphics[width=1.05\textwidth]{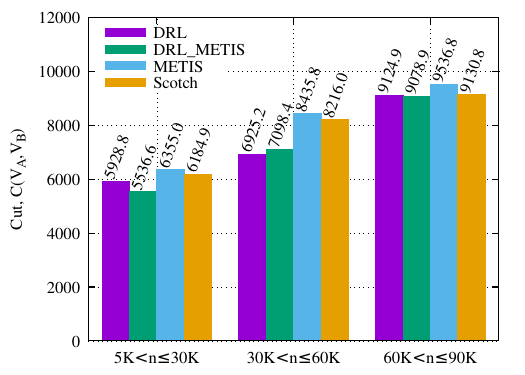}
      \caption{Edge cut}
      \label{fig:cuts_SuiteSparse}
   \end{subfigure}
   \begin{subfigure}[b]{0.32\textwidth}
      \centering
      \includegraphics[width=1.05\textwidth]{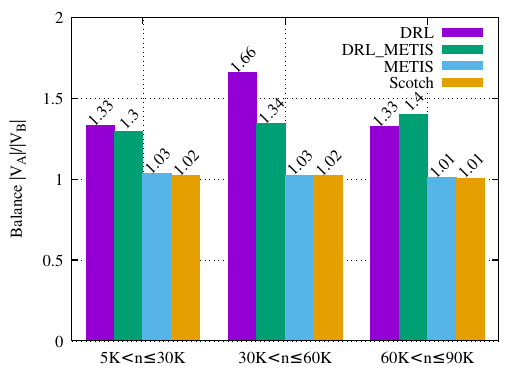}
      \caption{Partition balance}
      \label{fig:balance_SuiteSparse}
   \end{subfigure}
   \caption{Evaluation of the partitioners on the SuiteSparse dataset. Training was performed on graphs from the SuiteSparse collection with $100 < n \leq 5000$ nodes.}
   \label{fig:SuiteSparse_nc_c_balance}
\end{figure}

In order to see if the model is able to generalize to unseen graphs, we tested it on larger 2D/3D discretizations, with between $5,\!000$ and $90,\!000$ nodes (again, for testing we do not include coarser graphs). Figure \ref{fig:SuiteSparse_nc_c_balance} shows the normalized cut, the cut size and the balance for these graphs. The agent is able to generalize to these graphs, producing a lower cut and a lower normalized cut. The partitions are slightly unbalanced for graphs in the mid range of nodes.

% , on training set if we don't have enough data?}

\section{Finding a Minimal Vertex Separator\label{sec:vertex_separator}}

\begin{figure}
    \centering
    \includegraphics[scale=1.1]{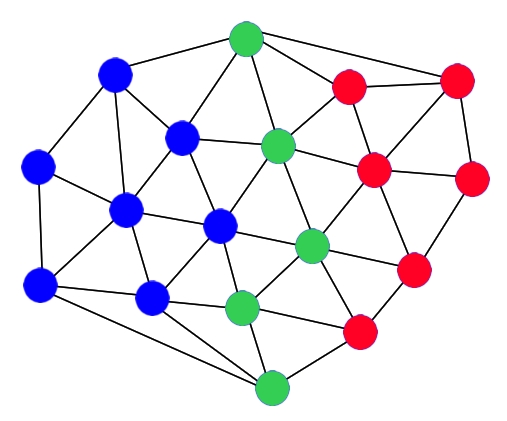}
    \caption{Example of a vertex separator. The vertices in the separator are depicted in green, while the other two partitions are depicted in red and blue. Note that the subgraph obtained by removing the vertices in the separator has exactly two connected components. In this case, the two components have approximately the same cardinality.}
    \label{fig:vertexSeparator}
\end{figure}

A vertex separator is a set of nodes that, if removed from the graph, would split the graph in two unconnected sub-graphs, see Figure \ref{fig:vertexSeparator}. Hence, the graph is partitioned in three subgraphs $G_A = (V_A,E_A)$, $G_B = (V_B,E_B)$ and $G_S = (V_S,E_S)$, such that $V = V_A \cup V_B \cup V_S$ and there are no edges connecting $V_A$ and $V_B$. The aim is to minimize $|V_S|$ while keeping the two other graphs $V_A$ and $V_B$ balanced. While the normalized cut (Eq. \ref{eq:normalized_cut}), which was used for graph bisection as discussed in Section \ref{sec:edge_separator}, is based on the volumes of the partitions (Eq. \ref{eq:volume}), we will now try to balance the cardinalities $|V_A|$ and $|V_B|$.  More precisely, we attempt to minimize
\begin{equation}
    \text{NS}(G) = |V_S| \left( \frac{1}{|V_A|} + \frac{1}{|V_B|} \right) \, , \label{eq:normalized_separator}
\end{equation}
a measure for the normalized separator. Note that a vertex separator can be computed using a number of heuristics~\cite{leiserson1987orderings,hendrickson1998improving}, or can be constructed from an edge separator using a minimum cover approach~\cite{pothen1990computing,duff1981algorithms}.

\begin{algorithm}
	\textbf{Input:} graph $G(V, E)$ \\
	\textbf{Output:} vertex separator $V_S$, unconnected components $V_A$ and $V_B$, such that $V = V_S \cup V_A \cup V_B$
	\begin{algorithmic}[1]
		\Function{vertex\_separator}{$G$}
            \If{$|V| < n_{\text{min}}$}
                \State \Return metis\_separator($G$) \label{line:vertex_recursion_stop}
            \EndIf
    		\State $G^C, I^C \gets $ coarsen($G$) \hfill \Comment{get coarse graph and interpolation info}
    		\State $V^C_S, V^C_A, V^C_B \gets $ vertex\_separator($G^C$) \hfill \Comment{recursive call using coarse graph}
    		\State $V_S, V_A, V_B \gets V^C_S(I^C), V^C_A(I^C), V^C_B(I^C)$ \hfill \Comment{interpolate $V_S$, $V_A$ and $V_B$ from coarse to fine}
    		\State $V^0_S, V^0_A, V^0_B \gets V_S, V_A, V_B$ \hfill \Comment{keep a copy}
    		\State $G^{\text{sub}} \gets $ k\_hop\_subgraph($G$, $V_S$, $k_{\text{hops}}$) \hfill \Comment{subgraph with all nodes at most $k_{\text{hops}}$ from $V_S$}
    		\State  \hfill \Comment{construct feature tensor}
	    	\State $F(v) \gets \left[ v \in V_A \, , v \in V_B \, , v \in V_S \, , v \in \partial G^{\text{sub}} \, , v \in V_S^{\min} \, , \sfrac{|V_A|}{|V|} \, , \sfrac{|V_B|}{|V|} \right], \,  \forall \, v \in G^{\text{sub}}$
    		\State $S \gets \text{NS}(V_S, V_A, V_B)$ \hfill \Comment{compute normalized separator,~Eq. \ref{eq:normalized_separator}}
    		% \State $A, R \gets [\,], [\,]$ \hfill \Comment{actions, cumulative rewards}
    		\For{$t \gets 1 $ to $2|V_S^0|$}
	    	    \State policy$_t$ $ \gets $ agent($G^{\text{sub}}, F$, $\{2, 3\}$) \hfill \Comment{forward evaluation of agent's neural network}
	    	    \State $a_t \gets $ argmax(policy$_t$) \hfill \Comment{pick action}
	    	    % \State $A \gets \left[A, \, a \right]$
	    	    \State apply\_action\_vertex\_separator($a_t, G, V_S, V_A, V_B$) \hfill \Comment{take an action, see~Algorithm \ref{alg:action_vertex_separator}}
	    	    \State  \hfill \Comment{update features}
	    	    \State $F(v) \gets \left[ v \in V_A \, , v \in V_B \, , v \in V_S \, , v \in \partial G^{\text{sub}} \, , v \in V_S^{\min} \, , \sfrac{|V_A|}{|V|} \, , \sfrac{|V_B|}{|V|} \right], \,  \forall \, v \in G^{\text{sub}}$
                \State $S_{\text{old}} \gets S$
	    	    \State $S \gets \text{NS}(V_S, V_A, V_B)$ \hfill \Comment{compute normalized separator,~Eq. \ref{eq:normalized_separator}}
	    	    \State $r_t \gets S_{\text{old}} - S$ \hfill \Comment{compute reward}
	    	\EndFor
	    	\State $V_S, V_A, V_B \gets V^0_S, V^0_A, V^0_B$
	    	\For{$t \gets 1$ to argmax($r$)}
	    	    \State apply\_action\_vertex\_separator($a_t, G, V_S, V_A, V_B$) \hfill \Comment{take an action, see~Algorithm \ref{alg:action_vertex_separator}}
	    	\EndFor
	    	\State \Return $V_S, V_A, V_B$
		\EndFunction
	\end{algorithmic}
	\caption{Computing a vertex separator using deep reinforcement learning. Shown here is only the evaluation phase.}
	\label{alg:vertex_separator}
\end{algorithm}

Algorithm \ref{alg:vertex_separator} computes, for a given graph $G(V, E)$, a vertex separator $V_S$ that approximately minimizes Eq. \ref{eq:normalized_separator}. Algorithm \ref{alg:vertex_separator} follows the same structure as Algorithm \ref{alg:edge_separator}, with some significant differences. Most importantly, the reward used in Algorithm \ref{alg:vertex_separator} is based on Eq. \ref{eq:normalized_separator}. In Line \ref{line:vertex_recursion_stop}, to stop the recursion, now METIS is called to compute a vertex separator instead of an edge separator. 

As before, a separator (now a vertex separator) is computed first on the coarser graph and then interpolated back to the finer graph. Then this separator is refined using deep reinforcement learning applied to a subgraph $G^{\text{sub}}$ containing all nodes within a small number of hops from the vertex separator. In every step of the deep reinforcement learning episode a single node $a_t$ is selected. Algorithm \ref{alg:action_vertex_separator} illustrates the action that is taken for a selected node $a_t$. If node $a_t$ is part of $V_A$, it is simply removed from partition $V_A$ and added to the separator $V_S$ (Line \ref{line:at_VA}). Likewise, if $a_t$ was in $V_B$, then node $a_t$ is removed from $V_B$ and added to $V_S$ (Line \ref{line:at_VB}). The partition volumes can be updated directly. If the selected node $a_t$ is part of the separator, we try to remove it from the separator $V_S$ and move it to one of the partitions $V_A$ or $V_B$. However, doing so could result in a state where $V_S$ is no longer a valid separator. This would be the case if after removing $a_t$ from the separator -- and moving it to either $V_A$ or $V_B$ -- there would be a direct edge between $V_A$ and $V_B$. However, by construction of the feature tensor and the agent's neural network, such a node will never be selected, as discussed in more detail below. Algorithm \ref{alg:action_vertex_separator} thus assumes that it is safe to remove $a_t$ from the separator. To decide whether to move $a_t$ from $V_S$ to $V_A$ or, instead to $V_B$, it is checked if $a_t$ was already connected to either $V_A$ or $V_B$. If there is an edge from $a_t$ to any node in $V_A$, then $a_t$ can be moved to $V_A$ (Line \ref{line:at_VStoVA}). Likewise, if $a_t$ was already connected to $V_B$ it can be moved to $V_B$ (Line \ref{line:at_VStoVB}). If $a_t$ has no edge to either $V_A$ or $V_B$, then $a_t$ is added to whichever one of $V_A$ or $V_B$ has the smallest cardinality (Line \ref{line:smallest_card}). Doing so will always result in a positive reward: the separator becomes smaller and the balance improves.

\begin{algorithm}
	\textbf{Input:} node $a$, graph $G(V, E)$, partitions $V_S$, $V_A$ and $V_B$ \\
	\textbf{Result:} node $a$ is moved to a different partition
	\begin{algorithmic}[1]
		\Procedure{apply\_action\_vertex\_separator}{$a, G(V, E), V_S, V_A, V_B$}
	        \If{$a \in V_A$}
	    	    \State $V_A, V_S \gets V_A \setminus a, V_S \cup a$ \hfill \Comment{move node $a$ from $V_A$ to $V_S$} \label{line:at_VA}
	        \ElsIf{$a \in V_B$}
	    	    \State $V_B, V_S \gets V_B \setminus a, V_S \cup a$ \hfill \Comment{move node $a$ from $V_B$ to $V_S$} \label{line:at_VB}
	    	\ElsIf{$a \in V_S$}
	    	    \State $V_S \gets V_S \setminus a$
	    	    \If{$\exists \, e_{a,v} \in E: v \in V_A$} \hfill \Comment{node $a$ is in $V_S$ and connected to $V_A$}
	    	        \State $V_A \gets V_A \cup a$ \hfill \Comment{move $a$ to $V_A$}
	    	    \ElsIf{$\exists \, e_{a,v} \in E: v \in V_B$} \hfill \Comment{node $a$ is in $V_S$ and connected to $V_B$} \label{line:at_VStoVA}
	    	        \State $V_B \gets V_B \cup a$ \hfill \Comment{move $a$ to $V_B$} \label{line:at_VStoVB}
	    	    \ElsIf{$|V_A| \leq |V_B|$} \label{line:smallest_card} \hfill \Comment{$a$ is not connected to $V_A$ or $V_B$}
	    	    \State $V_A \gets V_A \cup a$ \hfill \Comment{moving $a$ to $V_A$ improves the balance}
	    	    \Else \hfill \Comment{$|V_B| < |V_A|$}
	    	        \State $V_B \gets V_B \cup a$ \hfill \Comment{moving $a$ to $V_B$ improves the balance}
	    	    \EndIf
	    	\EndIf
		\EndProcedure
	\end{algorithmic}
	\caption{Apply an action to a vertex separator. This is used in Algorithm \ref{alg:vertex_separator} to compute a minimal vertex separator. This assumes node $a$ is not an essential node for the separator, i.e., node $a$ is not connected to both $V_A$ and $V_B$.}
	\label{alg:action_vertex_separator}
\end{algorithm}

Since we assume that the vertex separator computed at the coarser graph is of high quality, only a small number of refinement steps will be required. However, since moving the vertex separator by one node requires two actions -- first adding a node to the separator, then removing one -- the length of an episode is set to $2|V_S^0|$, i.e., twice the size of the interpolated separator.

For brevity, the training is omitted from Algorithm \ref{alg:vertex_separator}, since it can be added similarly to Algorithm \ref{alg:edge_separator}.

\paragraph{The Feature Tensor}
The feature tensor used in Algorithm \ref{alg:vertex_separator} is similar to the one used in Algorithm \ref{alg:edge_separator} to compute an edge separator. However, there are two additional features. One extra features comes from the fact that there are now three partitions instead of two, so the one-hot encoding of the partitioning is now $\left[ 1, \, 0, \, 0 \right]$, $\left[ 0, \, 1, \, 0 \right]$ and $\left[ 0, \, 0, \, 1 \right]$ for $V_A$, $V_B$ and $V_S$ respectively. The next feature again denotes whether a node is part of $\partial G^{\text{sub}}$, in which case it should not be selected as an action. The next features is a binary feature that is set to $1$ for the nodes that are essential to the separator, i.e., removing this node from the separator would lead to an invalid state. In Algorithm \ref{alg:vertex_separator} this set of nodes is denoted as $V_S^{\text{min}}$, and these are all nodes $v \in V_S : \exists \, e_{v,i}, e_{v,j} \text{ with } i \in V_A, \, j \in V_B$, i.e., node $v$ is in the separator and is connected to both $V_A$ and $V_B$. The final two features are $|V_A|/|V|$ and $|V_B|/|V|$.

\subsection{Experimental Evaluation}

\paragraph{Delaunay Triangulations}
We train the algorithm on Delaunay triangulations with between $100$ and $1,\!000$ nodes. As for the normalized cut, all intermediate coarsened graphs are included in the dataset, for a total of $\sim 10,\!000$ training graphs.

\begin{figure}
    \centering
    \begin{subfigure}[b]{0.32\textwidth}
      \centering
      \includegraphics[width=\textwidth]{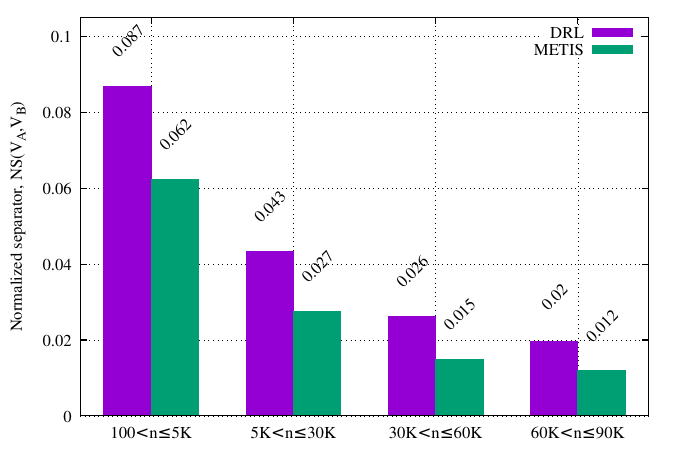}
    \caption{Delaunay}
    \label{fig:vertex_Delaunay}
    \end{subfigure}
   \begin{subfigure}[b]{0.32\textwidth}
      \centering
      \includegraphics[width=\textwidth]{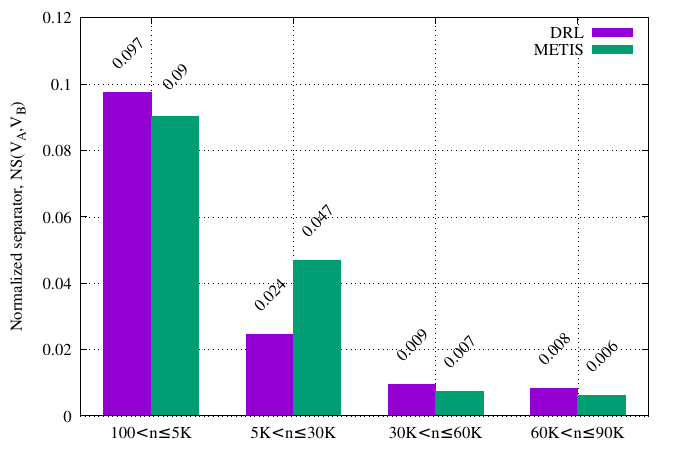}
      \caption{Hole6}
      \label{fig:vertex_Hole6}
   \end{subfigure}
   \begin{subfigure}[b]{0.32\textwidth}
      \centering
      \includegraphics[width=\textwidth]{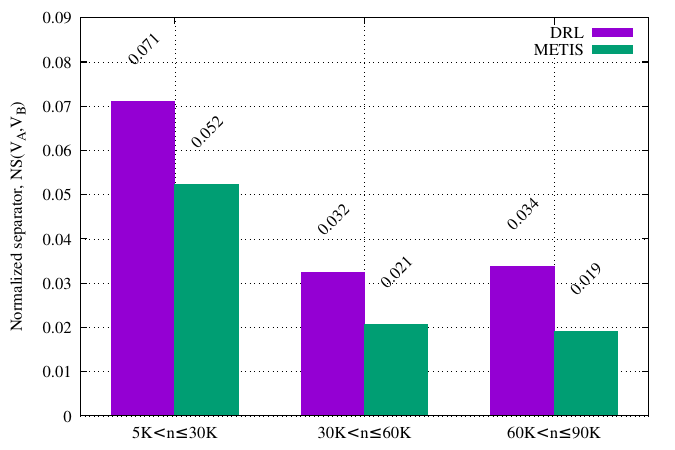}
      \caption{SuiteSparse}
      \label{fig:vertex_suitesparse}
   \end{subfigure}
   \caption{Comparison of the normalized separator, Eq. \ref{eq:normalized_separator}, using the proposed Algorithm \ref{alg:vertex_separator}, and METIS.}
   \label{fig:vertex_norm_sep}
\end{figure}

Figure \ref{fig:vertex_Delaunay} shows the normalized separator, Eq. \ref{eq:normalized_separator}, for the Delaunay testing sets corresponding to the parameters as given in Table \ref{tab:train_test_edge_params}. Figure \ref{fig:vertex_Hole6} shows the normalized separator for the Hole6 dataset. 
% The balance is defined as $\max\{|A|/|B|, |B|/|A|\}$ in this case. 
For the DRL experiments, METIS is used to find a vertex separator at the coarsest level. Our model produces slightly larger normalized separators than METIS, but the difference is small in all the different tests. Also, the algorithm is able to generalize extremely well also to the unseen triangulations of the Hole6 dataset, and it outperforms METIS for graphs having between $5,\!000$ and $30,\!000$ vertices.

\paragraph{Sparse matrices from the SuiteSparse matrix collection} As in the edge cut case, we train our agent on some 2D/3D discretizations from the SuiteSparse matrix collection. The dataset is the same as the one discussed in Section \ref{sec:edge_experiments} for the SuiteSparse experiments, and so is the structure of the deep neural network. 

Testing was done on 101 2D/3D discretizations from the SuiteSparse dataset with number of vertices between $5,\!000$ and $90,\!000$. Figure \ref{fig:vertex_suitesparse} shows the results for this dataset. Also in this case, the normalized separator values found with DRL are slightly higher than the ones from METIS, but overall the results are good. Indeed, the testing dataset includes larger graphs with a wide variety of sparsity patterns, many of them not included in the training dataset.

% \paragraph{Graded L-Shaped Domain}
% \todo[inline]{Alice: only graded L or also hole*?}
% \todo[inline]{show that the method generalizes}

\section{Nested Dissection Sparse Matrix Ordering\label{sec:nested_dissection}}

\begin{figure}
    \centering
    \includegraphics[width=\textwidth]{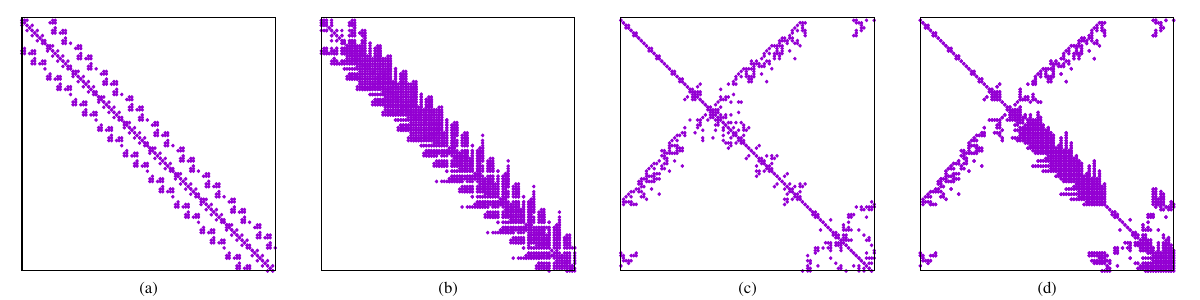}
    \caption{Nested dissection ordering of the $100\times 100$ sparse matrix $A = \text{nos4}$ from the SuiteSparse matrix collection. It is structurally symmetric, so its sparsity pattern can be represented by a graph with $100$ nodes and $594$ edges. (a) The sparsity pattern of the original matrix $A$, with $594$ number of non-zero entries (nnz). (b) The sparsity pattern of $L$ and $U$, the triangular factorization of $A = LU$, with nnz $ = 1417$. (c) The matrix symmetrically permuted using the minimum fill-reducing ordering: $P^T A P$, with nnz $=594$. (d) Sparsity pattern of the $LU$ factorization of the ordered matrix $P^TAP = LU$, with nnz $=1160$.}
    \label{fig:ND}
\end{figure}

The nested dissection algorithm is a heuristic used to order the rows and columns of a sparse matrix before applying Gaussian elimination to the matrix. When computing an LU decomposition of a sparse matrix $A$ as $A = LU$, the
%sparse
factors $L$ (lower triangular) and $U$ (upper triangular) are typically less sparse than the original matrix $A$. The extra nonzeros which are introduced in the factors are known as the fill-in. However, reordering the matrix before performing the numerical factorization can greatly reduce this fill-in, and  nested dissection is known to produce orderings that drastically reduce the fill. This is illustrated in Figure \ref{fig:ND} for the matrix \texttt{nos4}\footnote{https://sparse.tamu.edu/HB/nos4} from the SuiteSparse matrix collection. 
\begin{comment}
This matrix has 468 rows and columns and 5172 nonzeros elements. It is structurally symmetric, so its sparsity pattern can be represented by a graph with 468 nodes and 5172 edges. Figure \ref{fig:ND}(a) shows the sparsity pattern of the original matrix, (b) shows the sparse factors $L$ and $U$ of the original matrix, (c) is the matrix permuted with a minimum degree fill-reducing ordering, and finally (d) shows the $L$ and $U$ factors of the reordered matrix. Reordering the matrix reduces the nonzeros in the LU factors from $54,\!882$ to $36,\!872$.
\end{comment}

% nnz(A): 5172
% natural ordering nnz(L) + nnz(U): 54882
% permuted nnz(L) + nnz(U): 36872

Nested dissection orders the rows and columns of a sparse matrix $A$ as follows. Consider the graph $G$ corresponding to the sparsity pattern of $A$ if $A$ is symmetric, or to $A^T + A$ otherwise. Find a minimal vertex separator $V_S$ of $G$, that splits $G$ into two unconnected sets of nodes $V_A$ and $V_B$. Then first order the matrix rows and columns corresponding to the nodes in $V_A$ using nested dissection (recursively), next, number the rows and columns corresponding to the nodes in $V_B$ using nested dissection, and lastly, order the rows and columns corresponding the nodes in the vertex separator $V_S$.

\begin{algorithm}
	\textbf{Input:} graph $G(V, E)$ corresponding to the sparsity pattern of a symmetric matrix $A$, or to $A^T + A$ \\
	\textbf{Output:} permutation vector $p$
	\begin{algorithmic}[1]
	    \Function{nested\_dissection}{$G$}
            \If{$|V| < n_{\text{min}}$}
                \State \Return AMD($G$) \hfill \Comment{end recursion by calling AMD ordering~\cite{amestoy1996approximate}} \label{line:nd_stop_recursion}
            \EndIf
            \State $V_S, V_A, V_B \gets $ vertex\_separator($G$) \hfill \Comment{See Algorithm \ref{alg:vertex_separator}}
            % \State $G_A \gets $ subgraph($V_A, G$)
            % \State $G_B \gets $ subgraph($V_B, G$)
            \State $p_A \gets $ nested\_dissection(subgraph($V_A, G$)) \hfill \Comment{recursive call} \label{line:nd_rec1}
            \State $p_B \gets $ nested\_dissection(subgraph($V_B, G$)) \hfill \Comment{recursive call} \label{line:nd_rec2}
            \State \Return $\left[ V_A(i) \text{ for i in } p_A \right] + \left[ V_B(i) \text{ for i in } p_B \right] + V_S$
        \EndFunction
	\end{algorithmic}
	\caption{Nested dissection sparse matrix reordering, using deep reinforcement learning to find a vertex separator (Algorithm \ref{alg:vertex_separator}).}
	\label{alg:nested_dissection}
\end{algorithm}

Algorithm \ref{alg:nested_dissection} shows the nested dissection algorithm, using the recursive formulation for ease of notation. However, in the actual implementation this recursion is avoided using an implementation based on a stack data structure. Line \ref{line:nd_rec1} and Line \ref{line:nd_rec2} show the recursive calls. The recursion is stopped early, in Line \ref{line:nd_stop_recursion}, by calling a different ordering algorithm once the graph becomes smaller than the threshold $n_{\text{min}}$. Here we use the approximate minimum degree ordering (AMD)~\cite{amestoy1996approximate}, which is a different heuristic known to produce good ordering for small to medium sized problems, and for which highly efficient sequential codes are available. Algorithm \ref{alg:nested_dissection} returns a permutation vector $p$, corresponding to a permutation matrix $P$, which can be used to symmetrically permute the rows and the columns of the matrix.

\subsection{Experimental Evaluation}
We compare several sparse matrix ordering techniques on a number of sparse matrices by computing a factorization using the SuperLU sparse direct solver~\cite{li2005overview}, through the Scipy Python interface~\cite{2020SciPy-NMeth}.
The orderings considered are: 
\begin{itemize}  \setlength\itemsep{0em}
   % \item \textbf{NATURAL:} no reordering, use the matrix as provided
    \item \textbf{DRL\_ND:} nested dissection using Algorithm \ref{alg:nested_dissection}, computing the vertex separators using Algorithm \ref{alg:vertex_separator}.
    \item \textbf{METIS\_ND:} nested dissection using Algorithm \ref{alg:nested_dissection}, but computing the vertex separators using METIS.
    \item \textbf{METIS:} nested dissection implementation from METIS.
    \item \textbf{SCOTCH:} nested dissection implementation from SCOTCH.
    \item \textbf{COLAMD:} approximate minimum degree ordering applied to the graph of $A^TA$ (the default option for SuperLU).
\end{itemize}

In Algorithm \ref{alg:nested_dissection}, $n_{\text{min}}$ is set to $100$.

\begin{figure}
    \begin{subfigure}[b]{0.49\textwidth}
      \centering
      \includegraphics[width=\textwidth]{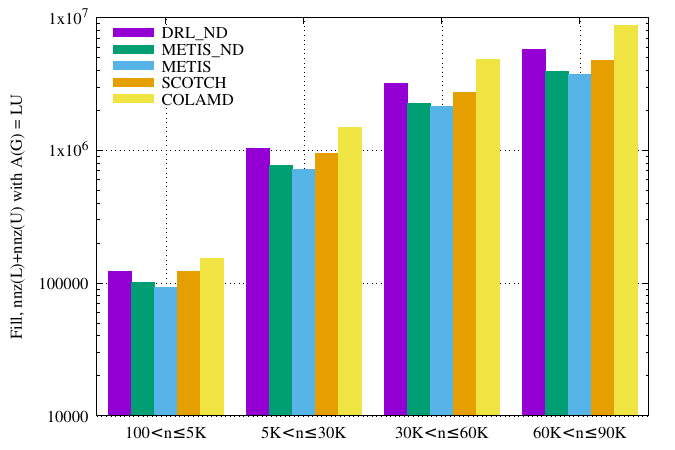}
      \caption{Delaunay graphs}
      \label{fig:fill_Delaunay}
    \end{subfigure}
    \begin{subfigure}[b]{0.49\textwidth}
      \centering
      \includegraphics[width=\textwidth]{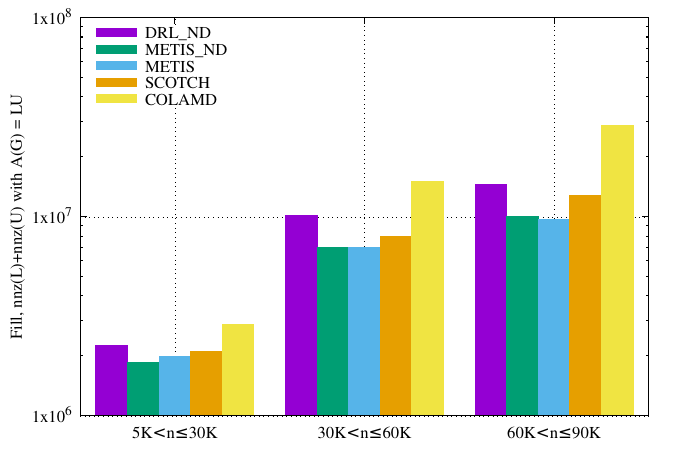}
      \caption{SuiteSparse graphs}
      \label{fig:fill_suitesparse}
    \end{subfigure}
    \caption{Comparison of the fill in the triangular factors $L$ and $U$ of a matrix $A$ corresponding to (Figure \ref{fig:fill_Delaunay}) the adjacency matrix (and a diagonal shift) of several Delaunay graphs or (Figure \ref{fig:fill_suitesparse}) several Suitesparse matrices. For DRL\_ND, the sparse matrix is reordered using Algorithm \ref{alg:nested_dissection} with Algorithm \ref{alg:vertex_separator}  to find the vertex separators. For METIS\_ND, the permutation is computed using Algorithm \ref{alg:nested_dissection}, but with METIS to find the vertex separators. The METIS datapoints in this figure use the nested dissection algorithm as implemented in the METIS package.}
   \label{fig:fill_Delaunay_suitesparse}
\end{figure}

\paragraph{Delaunay Triangulations}
Figure \ref{fig:fill_Delaunay} collects the results for the fill on $100$ Delaunay triangulations having between $100$ and $90,\!000$ vertices with the different orderings. Our model (DRL\_ND) always outperforms minimum degree (COLAMD) and it gives results similar results to the SCOTCH orderings, while performing slightly worse than METIS and nested dissection (Algorithm \ref{alg:nested_dissection}) with METIS to find the vertex separators (METIS\_ND). This shows that the model is able to generalize well to larger and unseen graphs.

\paragraph{Sparse Matrices from the SuiteSparse Collection}

Figure \ref{fig:fill_suitesparse} shows the results for the fill on $101$ 2D/3D discretizations from the SuiteSparse matrix collection, with number of vertices between $5,\!000$ and $90,\!000$. We notice that our proposed algorithm (DRL\_ND) performs similarly to SCOTCH and outperforms minimum degree (COLAMD) in every class of graphs. The fill with DRL turns out to be a little higher than the one obtained with METIS and METIS\_ND. Recall that this testing dataset includes larger graphs than the ones used for training and a wide variety of sparsity patterns, so the results show that the model generalizes well to larger graphs, and to graphs from different types of discretizations.

\section{Conclusions\label{sec:conclusion}}
We have presented a graph partitioning approach based on deep reinforcement learning, using a multilevel framework. We show both an edge partitioner, computing graph bisections, as well as a variant of the method which computes a vertex separator. 
% The edge separator approximately minimizes the normalized cut, while the vertex separator code minimizes the size of the separator while keeping the cardinalities of the partitions balanced. The 
We show that the graph partitioning and vertex separator codes, when trained on graphs of a certain type and with less than $5,\!000$ nodes, generalize well to graphs with many more nodes, as well as to different types of graphs. For instance, Figures \ref{fig:cuts_SuiteSparse} and \ref{fig:normalized_cuts_SuiteSparse} show that for graphs with up to $90,\!000$ nodes, the cut and the normalized cut  for the partitions from the proposed method, are very competitive with those computed using METIS and SCOTCH. %In fact, the average cut size is around $20\%$ smaller than for METIS and Scotch. 
Note, the graphs we used for this test are from the SuiteSparse dataset, which contains problems from a variety of applications, showing a wide range of sparsity patterns.

In Section \ref{sec:nested_dissection}, the vertex separator code is used recursively to construct a nested dissection ordering, which we then evaluate in the sparse solver SuperLU. We show that the resulting sparse matrix ordering effectively reduces the fill-in, and does this more so than the approximate minimum degree ordering. The quality of the ordering is comparable to the ordering produced by SCOTCH and only slightly worse than the nested dissection ordering from METIS. We believe further tuning of the neural network, and training on more and larger graphs, could further improve these results. 

From our complexity analysis in Section \ref{sec:edge_complexity}, we believe that the presented approach can also perform well, competitively with the state-of-the-art codes METIS and SCOTCH. As a possible improvement to the runtime of the algorithms, we could consider selecting multiple actions at once, effectively reducing the number of forward evaluations of the neural network. Furthermore, we plan to port the method from Python to compiled C++, to gain additional computation efficacy. %This should be relatively straightforward, especially since the neural network architecture, relying on the SAGE graph convolutional layers, is relatively simple.

% \todo[inline]{stress the fact that the results are good and that it generalizes well, but also that at the moment the performance is not excellent, but it can be improved. I agree about the deployment, and also our model should be easily integrated. To reduce cost, we can flip multiple nodes at once, if they are for enough.}

\section*{Acknowledgements}
This work was supported by the Laboratory Directed Research and Development Program of Lawrence Berkeley National Laboratory under U.S. Department of Energy Contract No. DE-AC02-05CH11231.

\bibliographystyle{plain}
\bibliography{refs}

\end{document}